\documentclass[journal]{IEEEtran}

\usepackage{xcolor,soul,framed} 

\colorlet{shadecolor}{yellow}
\usepackage[pdftex]{graphicx}
\graphicspath{{../pdf/}{../jpeg/}}
\DeclareGraphicsExtensions{.pdf,.jpeg,.png}

\usepackage[cmex10]{amsmath}
\usepackage{array}
\usepackage{mdwmath}
\usepackage{mdwtab}
\usepackage{eqparbox}
\usepackage{url}

\usepackage[utf8]{inputenc}
\usepackage[T1]{fontenc}

\usepackage[cmex10]{amsmath}

\usepackage{amsfonts}
\usepackage{algorithmic}
\usepackage{algorithm}
\usepackage{array}
\usepackage{textcomp}
\usepackage{stfloats}
\usepackage{url}
\usepackage{verbatim}
\usepackage{graphicx}
\usepackage{xcolor}
\usepackage{soul}
\usepackage{svg}
\usepackage{multirow}
\usepackage{hyperref}
\usepackage{soul}
\usepackage{xcolor}
\usepackage[caption=false]{subfig}
\usepackage{booktabs}

\usepackage{amssymb}

\begin{document}


\title{Thermal Vision Soil Assessment for Robotic Navigation on a Multipurpose Environmental Chamber under Simulated Mars Conditions}

\author{{Raúl Castilla-Arquillo, Anthony Mandow, Carlos J. Pérez-del-Pulgar-Mancebo, César Álvarez-Llamas, \\ José M. Vadillo, and Javier Laserna}

\thanks{R. Castilla-Arquillo, A. Mandow and C. J. Pérez-del-Pulgar-Mancebo are with the Department of Systems Engineering and Automation, University of Málaga, Andalucía Tech, 29070 Málaga, Spain.}
\thanks{C.Álvarez-Llamas, J. M. Vadillo and J. Laserna are with UMALASERLAB, Department of Analytical Chemistry, University of Málaga, 29010 Málaga, Spain.}}


\maketitle

\begin{abstract}
Soil assessment is important for mobile robot planning and navigation in both natural and planetary environments. Thermal behaviour of soils under sunlight, as observed through remote sensors such as Long-Wave Infrared cameras, allows for the inference of soil characteristics. However, this behaviour is greatly affected by the low atmospheric pressures of planets such as Mars. Measurement systems are needed that can simulate both the behaviour and realistic measurement limitations under the conditions encountered during planetary exploration. This article proposes a remote thermal measurement system based on multipurpose environmental chambers to physically simulate soils thermal behaviour over diurnal cycles under planetary conditions of pressure and atmospheric composition. To validate the proposed system, we analyze the thermal behaviour of four soil samples of different granularity by comparing replicated Martian surface conditions and their Earth's diurnal cycle equivalent. The obtained results indicate a correlation between granularity and thermal inertia that is consistent with available Mars surface measurements recorded by rovers. The proposed system enables the generation of datasets to train machine learning algorithms focused on on-site terrain segmentation based on soils characteristics using thermal vision with applicability in future planetary exploration missions. The dataset resulting from the experiments has been made available for the scientific community.
\end{abstract}

\begin{IEEEkeywords}
Soil Assessment, Thermal Inertia, Thermal Vision, Multipurpose Environmental Chamber.
\end{IEEEkeywords}


\section{Introduction}

\IEEEPARstart{R}{emote} assessment of soils characteristics can be crucial for the safety and efficiency of a broad range of tasks related to planetary mobile robot navigation such as odometry, environment mapping or energetic consumption~\cite{wong2022theory}. Soil assessment is useful to prevent slipping, skidding and getting entrapped on granular soils, which led to delay and significant mobility difficulties in the Curiosity and Spirit rover missions~\cite{arvidson2017mars},~\cite{gonzalez2018slippage},~\cite{feng2022instrumented}. 

In general, onboard mobile robot sensors such as RGB stereo cameras or 3D laser scanners~\cite{guastella2020learning} can be used to infer soil characteristics such as  roughness and slope~\cite{nampoothiri2021recent}. However, these measurements are limited to the surface layer, so relevant subsurface properties for traversability such as soil cohesion or internal friction cannot not be assessed. Alternatively, infrared data has been useful for terrain classification on Mars~\cite{putzig2007apparent}. In this sense, onboard remote sensors such as thermopiles and thermal cameras can provide relevant data to infer subsurface properties from thermal behaviour~\cite{chhaniyara2012terrain}. 

Thermopiles are being used in the  Curiosity and Perseverance rovers~\cite{gomez2012rems},~\cite{perez2018thermal} to perform on-site measurements of Martian surface thermal behaviour. Furthermore, in the future Martian Moons eXploration (MMX) mission, a rover will be equipped with thermophiles to infer Phobos' composition from its thermal inertia~\cite{michel2022mmx}. 
Nevertheless, thermal cameras offer significantly higher resolution, which can be advantageous for assessment and segmentation of heterogeneous soils. For instance, thermal images can be processed to infer soil traversability from measured thermal diffusivity~\cite{cunningham2015predicting} or thermal inertia~\cite{cunningham2015terrain},~\cite{gonzalez2017thermal}.

\begin{figure}[t]
    \centering
    \includegraphics[width=0.45\textwidth]{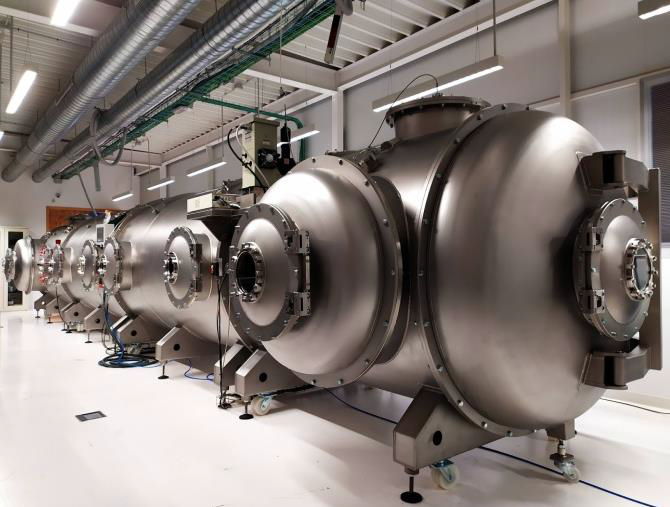}
    \caption{UMA-Laserlab Mars Environment Chamber used in the experiments.}
    \label{fig:panoramic-martian-chamber}
\end{figure}

Thermal imagery is suitable for training neural networks to classify soils based on their thermal behaviour~\cite{iwashita2020virtualIR}. In fact,  there is a growing interest in image-based machine learning for navigation and terrain classification for planetary rovers \cite{Rothrock2016},~\cite{Mandrake2022TrustedIA}.
In particular, thermal inertia measurements have been used to train slippage models on rovers~\cite{cunningham2019marsslip} and to improve their autonomy on machine learning systems~\cite{ono2020maars}. Nonetheless, machine learning approaches are limited by an insufficient amount of representative data, given the difficulty and expense of planetary imaging~\cite{Nagle-Mcnaughton2022},~\cite{Atha2022SPOC}. Besides, the thermal behaviour obtained in experiments on Earth is often different from the behaviour on planets such as Mars, which limits the applicability of the machine learning models~\cite{cunningham2019marsslip}. Therefore, measurement systems are needed to obtain experimental results on Earth that are representative of planetary conditions.

The ability to replicate  conditions representative of a real scientific mission on other planets is important in experiments for thermal inertia estimations, which are very dependent on pressure~\cite{putzig2006thermal}. In this sense, Multipurpose Environmental Chambers (MECs) can operate under representative conditions of temperature and pressure found in other planets such as Mars~\cite{vakkada2020space},~\cite{wu2021mars}. 

This article proposes a MEC-based remote thermal measurement system to physically simulate soils thermal behaviour over diurnal cycles under planetary conditions of pressure and atmospheric composition. To validate the proposed system, we analyzed the thermal behaviour of four soil samples of different granularity by comparing replicated Martian surface conditions and their Earth's diurnal cycle equivalent. The experiments were performed on the UMA-Laserlab Multipurpose Thermal Vacuum Chamber (see Fig.~\ref{fig:panoramic-martian-chamber}), which can replicate the atmospheric $CO_{2}$ composition of Mars. Results indicate a correlation between granularity and thermal inertia that is consistent with available Mars surface measurements recorded by rovers. The proposed system enables the generation of datasets to train machine learning algorithms focused on on-site terrain segmentation based on soils characteristics using thermal vision with applicability in future planetary exploration missions. 


This article is organized as follows. Section~\ref{sec:thermal-inertia} reviews methods to estimate the thermal inertia. Section~\ref{sec:framework} presents the proposed MEC-based remote thermal measurement system. Section~\ref{sec:experimental-setup} describes the experimental setup. Section~\ref{sec:experiments} evaluates the proposed measurement system. Finally, Section~\ref{sec:conclusions} offers conclusions and provides an insight on future works.

\section{Thermal inertia}
\label{sec:thermal-inertia}

This section reviews thermal inertia concepts, the use of the thermal diffusion equation to model the Martian surface thermal behaviour, and two methods to estimate thermal inertia based on surface temperature gradients. 

\subsection{Definition and pressure dependence}
Thermal inertia, $I$, is defined as follows:
\begin{equation}
I = \sqrt{k \rho c} ,
\end{equation}
where $k$ is the bulk thermal conductivity, $p$ is the bulk density and $c$ is the soil specific heat capacity. Thermal inertia is the property of a material that affects the resistance of a soil to change its temperature. A higher thermal inertia value means a slower heating of the soil. Thermal conductivity is the parameter which mainly influences thermal inertia, which is affected by three different heat transfer mechanisms~\cite{putzig2006thermal}:

\begin{equation}
k = k_{r} + k_{c} + k_{g},
\end{equation}
where $k_{r}$ is the transfer across pore spaces, $k_{c}$ is the conduction between grains contact areas, and $k_{g}$ is the conduction of the gas which fills the pores between grains. Pressure greatly determines which term acquires the most relevance. Gas conduction ($k_{g}$) dominates at pressures between $0.1\:mbar$ and $1000\:mbar$, where there is a near-linear relationship between particle size and thermal conductivity for granular soils~\cite{presley1997effect},~\cite{masamune1963thermal}. In this case, loose granular soils have lower thermal inertia than compacted rocky soils~\cite{jakosky1986thermal}. However, the relationship is not so strong at pressures higher than $1000\:mbar$. Thus, it is easier to estimate the soils characteristics based on thermal inertia at Martian pressure than at Earth pressure.


\subsection{Martian surface behaviour}
Martian surface thermal behaviour can be expressed as a boundary condition on the thermal diffusion equation derived from its surface energy budget:

\begin{equation}
  G = - I \sqrt{\frac{\pi}{P}} \left.\frac{\partial T}{\partial Z'}\right\vert_{Z'=0} = (1 - A) R_{sw} - \epsilon \sigma_{B} T^4_{s} + \epsilon R_{lw} - F_{CO_{2}} ,
  \label{eqn:mars-flux}
\end{equation}
where $G$ is the net heat flux expressed in $W/m^2$, $A$ is the albedo, $\sigma_{B}$ is the Stefan–Boltzmann constant, $R_{sw}$ is the down-welling shortwave (SW) radiation absorbed from the Sun, $R_{lw}$ is the down-welling longwave (LW) radiation emitted by the atmosphere and the Sun, $\epsilon$ is the thermal emissivity, $F_{CO_{2}}$ is the seasonal $CO_{2}$ condensation, P is the period of a diurnal cycle, $T_{s}$ is the surface temperature, and the term $\left.\frac{\partial T}{\partial Z'}\right\vert_{Z'=0}$ is the temperature gradient evaluated at the surface of the terrain, being $Z'$ the distance into the terrain normalized to the thermal skin depth. The sign convention is to use a positive sign when modeling the heating of the terrain and a negative sign when modeling its cooling.

The $F_{CO_{2}}$ term of~(\ref{eqn:mars-flux}) is negligible for Martian surfaces located from equatorial to mid-latitudes that present no frost. Moreover, the down-welling LW radiation is not considered, as previous on-ground measurements have shown it to be an order of magnitude smaller than the rest of terms~\cite{martinez2014surface}. Thus, the simplified equation for the soil thermal behaviour is:

\begin{equation}
  G = - I \sqrt{\frac{\pi}{P}} \left.\frac{\partial T}{\partial Z'}\right\vert_{Z'=0} =  (1 - A) R_{sw} - \epsilon \sigma_{B} T^4_{s},
\label{eqn:mars-simplified}
\end{equation}
where the soil thermal behaviour depends on the SW incident Sun's radiation, the thermal inertia and the surface radiative emission.

\subsection{Thermal inertia estimation}
Thermal inertia represents a complex combination of physical properties that are not directly measurable in practice, so simplified estimations based on surface temperature observations are required~\cite{wang2010methodTI}. In this work, we use the Apparent Thermal Inertia (ATI)~\cite{price1977thermal} and the method based on daily amplitude of surface soil heat flux and temperature in~\cite{wang2010methodTI}.

\begin{figure}[t]
    \centering
    \includegraphics[width=0.45\textwidth]{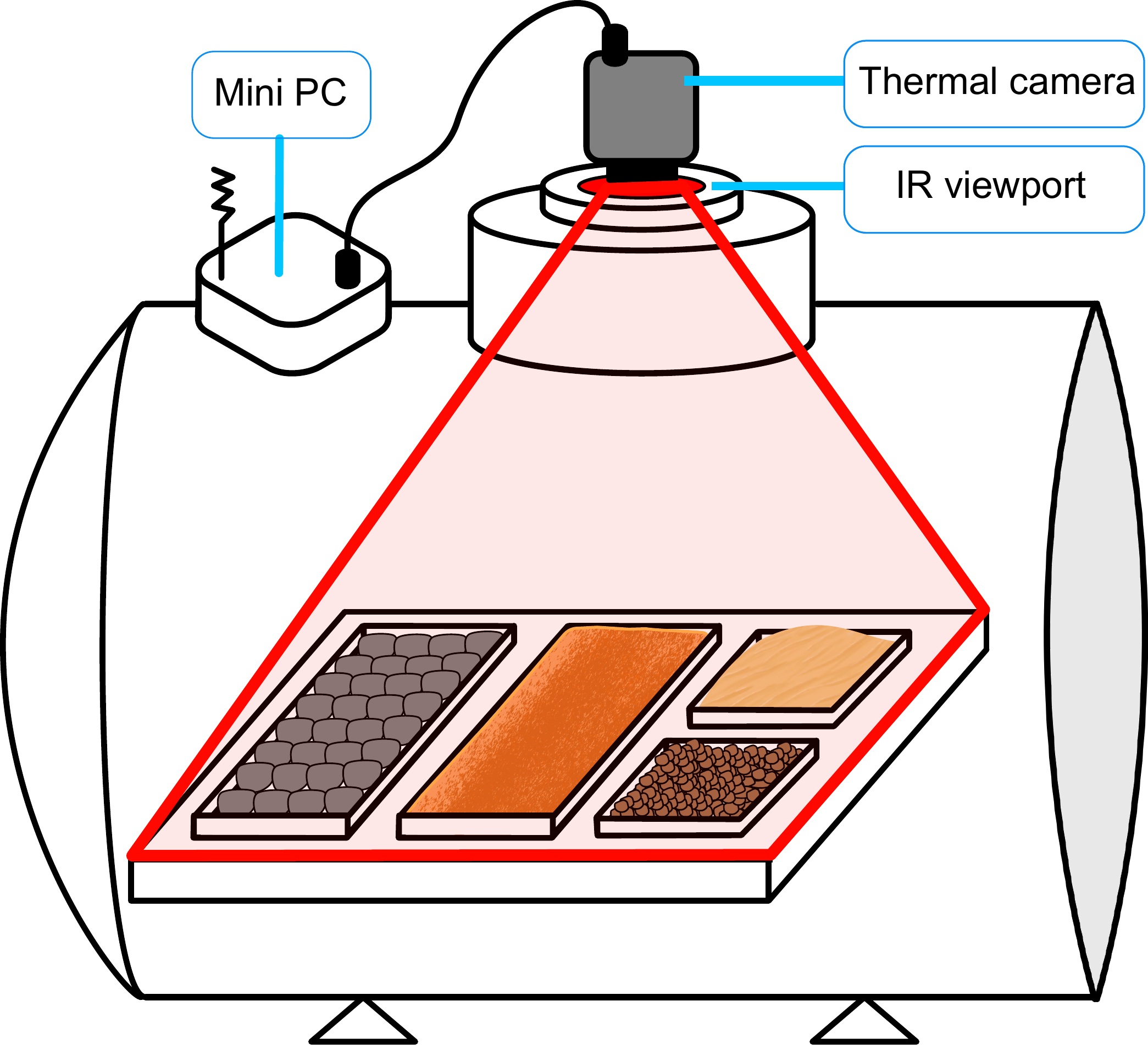}
    \caption{Schematic diagram of the proposed experimental MEC setup.}
    \label{fig:schematic-chamber}
\end{figure}

\begin{figure}[t]
    \centering
    \includegraphics[width=0.48\textwidth]{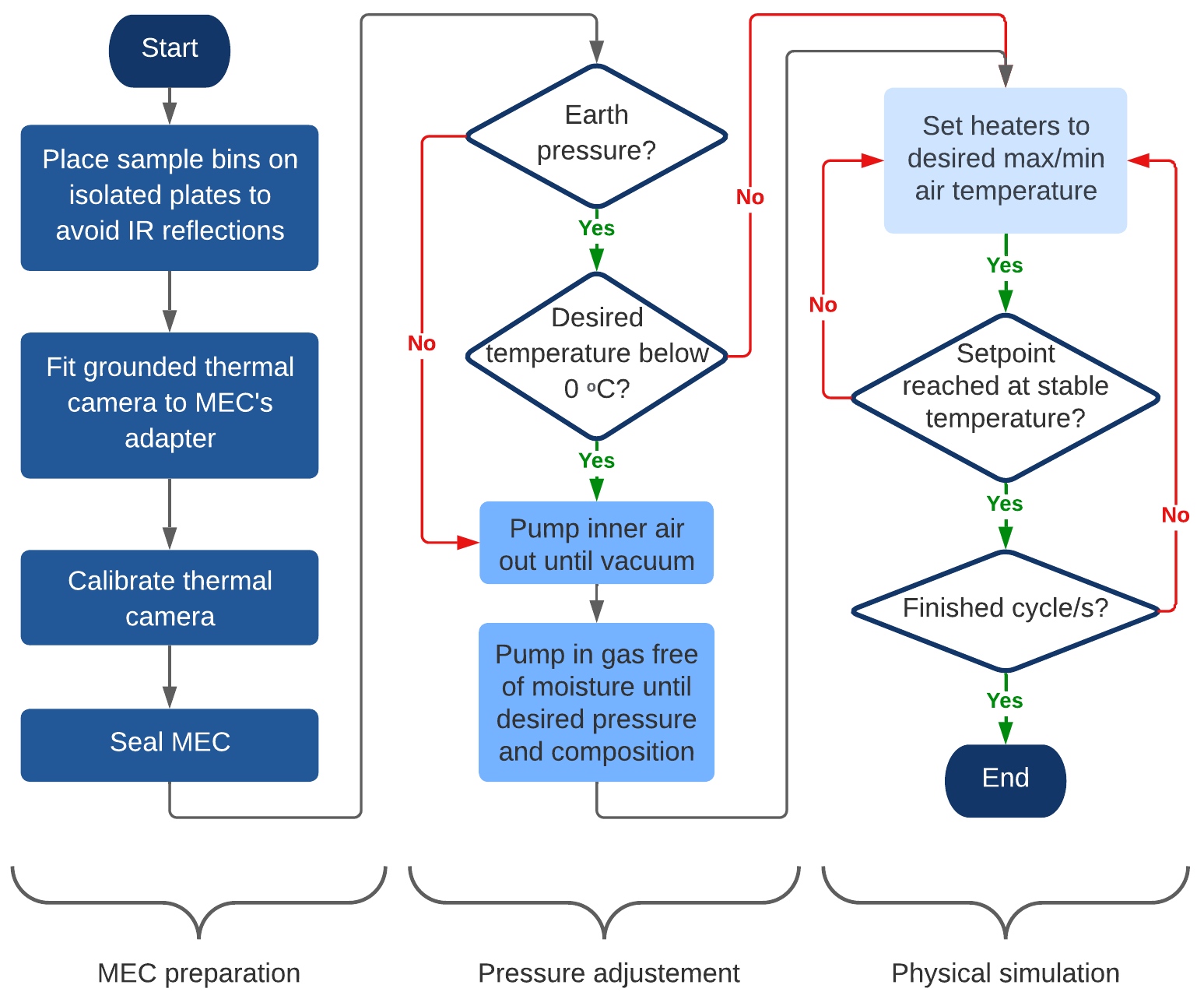}
    \caption{Flow chart of the proposed methodology.}
    \label{fig:flow-chart}
\end{figure}

ATI~\cite{price1977thermal} is a simple method to estimate the thermal inertia of an outdoors surface subjected to the Sun's heating. This estimation takes into account the diurnal temperature amplitude by measuring the minimum night and maximum day surface temperatures, $T_{min}$ and $T_{max}$, respectively. The formula to obtain the ATI of a surface is:

\begin{equation}
ATI = \frac{1-A}{\Delta T_{s}} , 
\label{eqn:ati}
\end{equation}
where $\Delta T_{s}= T_{max} - T_{min}$. The result can be multiplied by a 4186 coefficient to express ATI in thermal inertia units, $tiu \equiv\frac{Ws^{1/2}}{m^2K}$. Throughout this work, ATI will always be expressed in $tiu$. Even if ATI is widely used in the literature, this estimation does not consider the surface energy budget, among other limitations~\cite{price1985analysis}.

The alternative estimation proposed in~\cite{wang2010methodTI} considered a sinusoidal approximation of the Earth's net heat flux and surface temperatures for a diurnal period $P$. The same assumption can be applied to Mars' heat fluxes and temperatures~\cite{martinez2014surface}. Under this assumption, the thermal inertia of a given soil can be estimated as:

\begin{equation}
    I_{sin} = \frac{\Delta G_{s}}{\Delta T_{s} \sqrt{2\pi / P}} ,
\label{eqn:inertia-estimation}
\end{equation}
where the net heat flux expressed as $\Delta G_{s}= G_{max} - G_{min}$, being $G_{max}$ and $G_{min}$ the maximum and minimum values of the net heat flux, respectively.

\section{MEC-based remote thermal measurement system} \label{sec:framework}

In this section, we propose a MEC-based remote thermal measurement system to physically simulate soils thermal behaviour over diurnal cycles under planetary conditions of pressure and atmospheric composition. The system consists of a physical MEC-based configuration (see Fig.~\ref{fig:schematic-chamber}) and an experimental methodology (see Fig.~\ref{fig:flow-chart}). 

The proposed physical configuration  (see Fig.~\ref{fig:schematic-chamber}) allows to perform remote temperature measurements under the extreme conditions produced by the MEC. This configuration consists of an inner plate where the sample bins are placed, an external thermal camera connected to a Mini PC for data collection and an IR viewport. The viewport must allow the infrared range from $8\:\mu m$ to $14\:\mu m$ to pass through with minimal losses. Furthermore, it must withstand the pressure differential and temperatures reached by the MEC. 

The experimental methodology (see Fig.~\ref{fig:flow-chart}) consists of three sequential tasks: the preparation of the MEC setup; the adjustment of the inner pressure according to the environment to be replicated; and the physical simulation where the actuation profile is defined. In the preparation task, the sample bins are placed on the plate while care is taken to thermally insulate the plate surface, as it can produce IR reflections that can distort the measurements. The thermal camera is placed on the viewport and its housing is connected to ground to avoid the electrostatic charges produced by the MEC pumps. Next, the thermal camera is calibrated to provide precise measurements of the sample bins surfaces and, finally, the MEC gets sealed. In the pressure adjustment task, different procedures have to be performed depending on the pressure and temperature range of the experiments. The simulation can be started if the experiments are planned to be at earth pressure and temperature above $0^\circ C$. Otherwise, the air is pumped out until vacuum and then humidity-free air of the desired composition (i.e., $95\%$ of $CO_{2}$ for Mars) is pumped in. This process prevents the freezing of the air moisture from affecting the MEC internal systems. During the physical simulation part, temperatures are defined for the MEC heaters to obtain sinusoidal soil samples temperatures similar to those obtained in a diurnal cycle in reality.

In the physical simulation, the surface energy budget of each sample bin inside the MEC can be expressed in function of a radiative flux produced by the MEC heaters according to the following equation:

\begin{equation}
  G = - I \sqrt{\frac{\pi}{P}} \left.\frac{\partial T}{\partial Z'}\right\vert_{Z'=0} =  \epsilon \sigma_{B} T^4_{heater} - \epsilon \sigma_{B} T^4_{s_{i}},
\label{eqn:chamber-flux}
\end{equation}
where $T_{s_{i}}$ is the surface mean temperature of each soil and $T_{heater}$ is the MEC heaters temperature. Air natural convection is considered to be negligible as the MEC is an enclosed space with no wind. Inside the MEC, the radiation term $ \epsilon \sigma_{B} T^4_{heater}$ simulates the active Sun heating of the term $(1 - A) R_{sw}$ of~(\ref{eqn:mars-simplified}). 




\begin{figure}[t]
    \centering
    \subfloat[\label{sfig:germanium-window-cad}]{\includegraphics[height=0.18\textwidth]{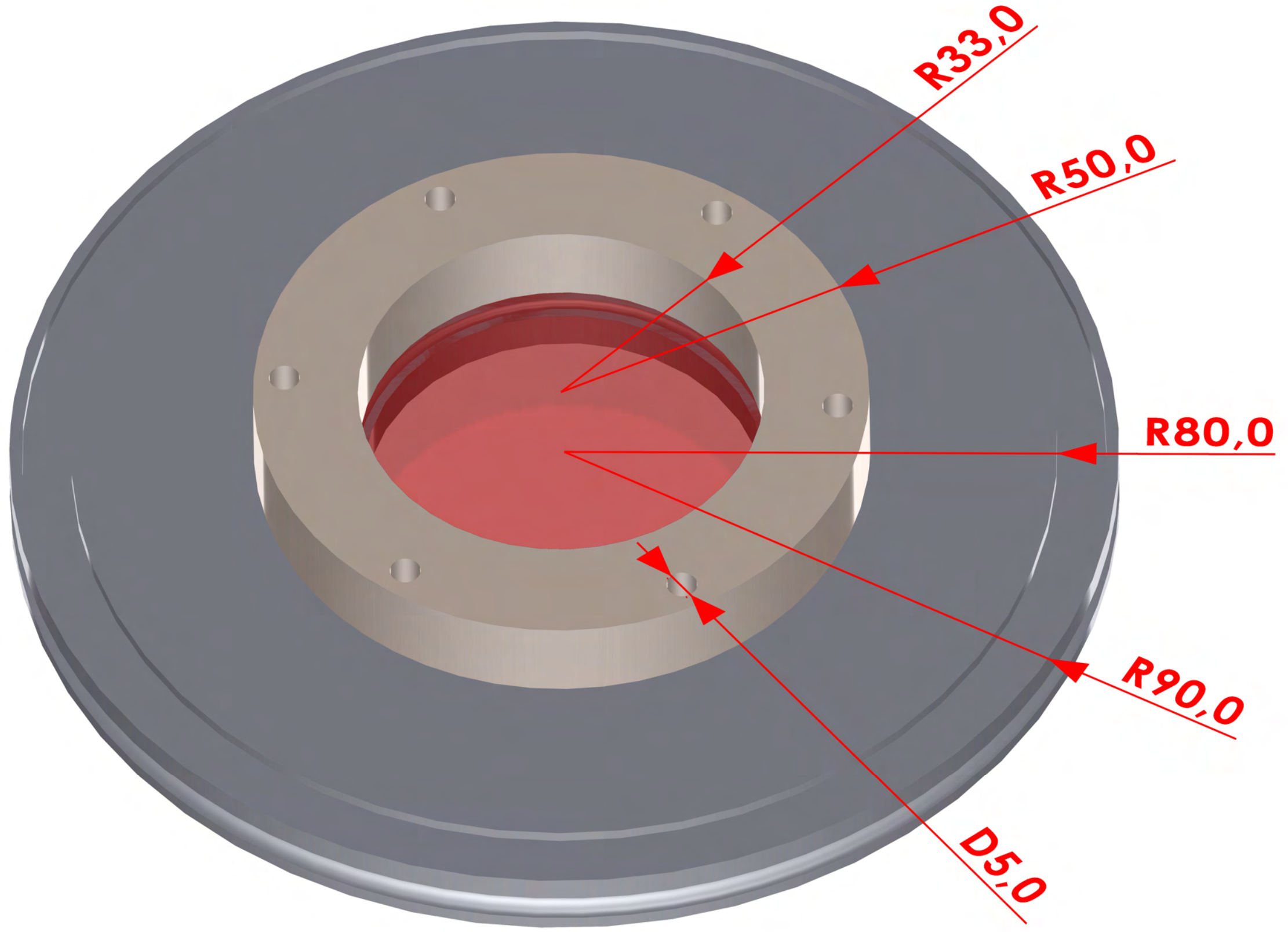}}
    \subfloat[\label{sfig:germanium-window}]{\includegraphics[height=0.18\textwidth]{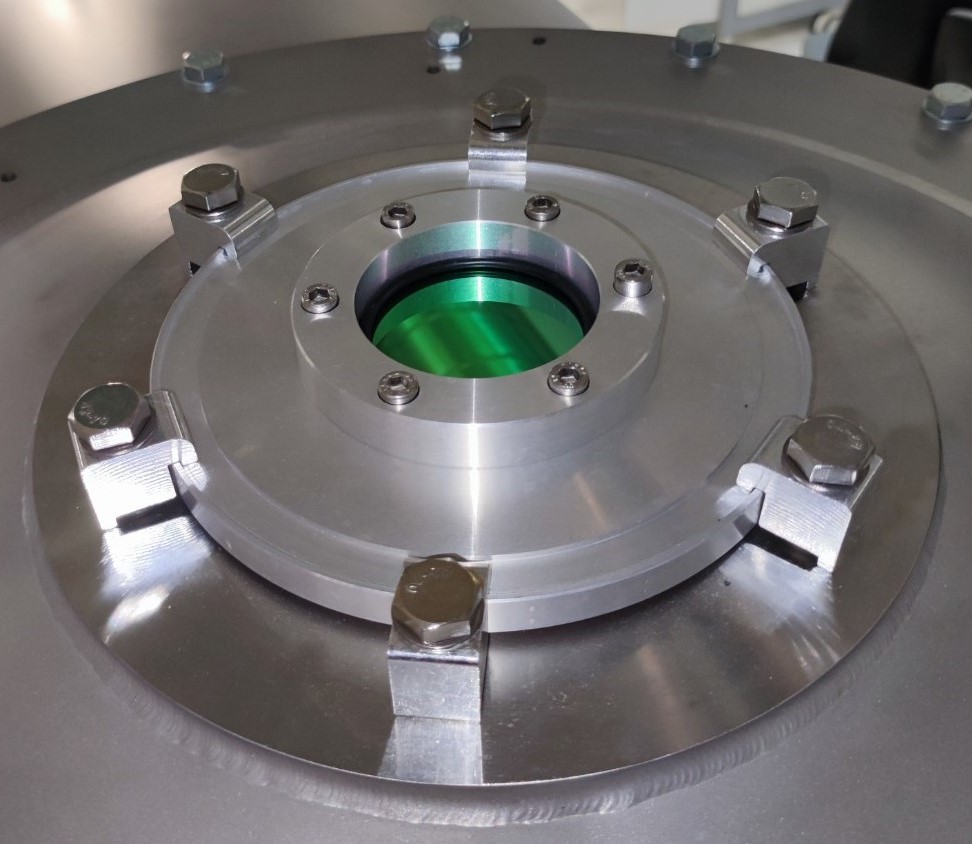}}
    \caption{a) 3D model of the viewport adapter (dimensions in millimeters) , b) Customized viewport adapter.}
    \label{ffig:martian-viewport}
\end{figure}

\begin{figure}[t]
    \centering
    \includegraphics[width=0.49\textwidth]{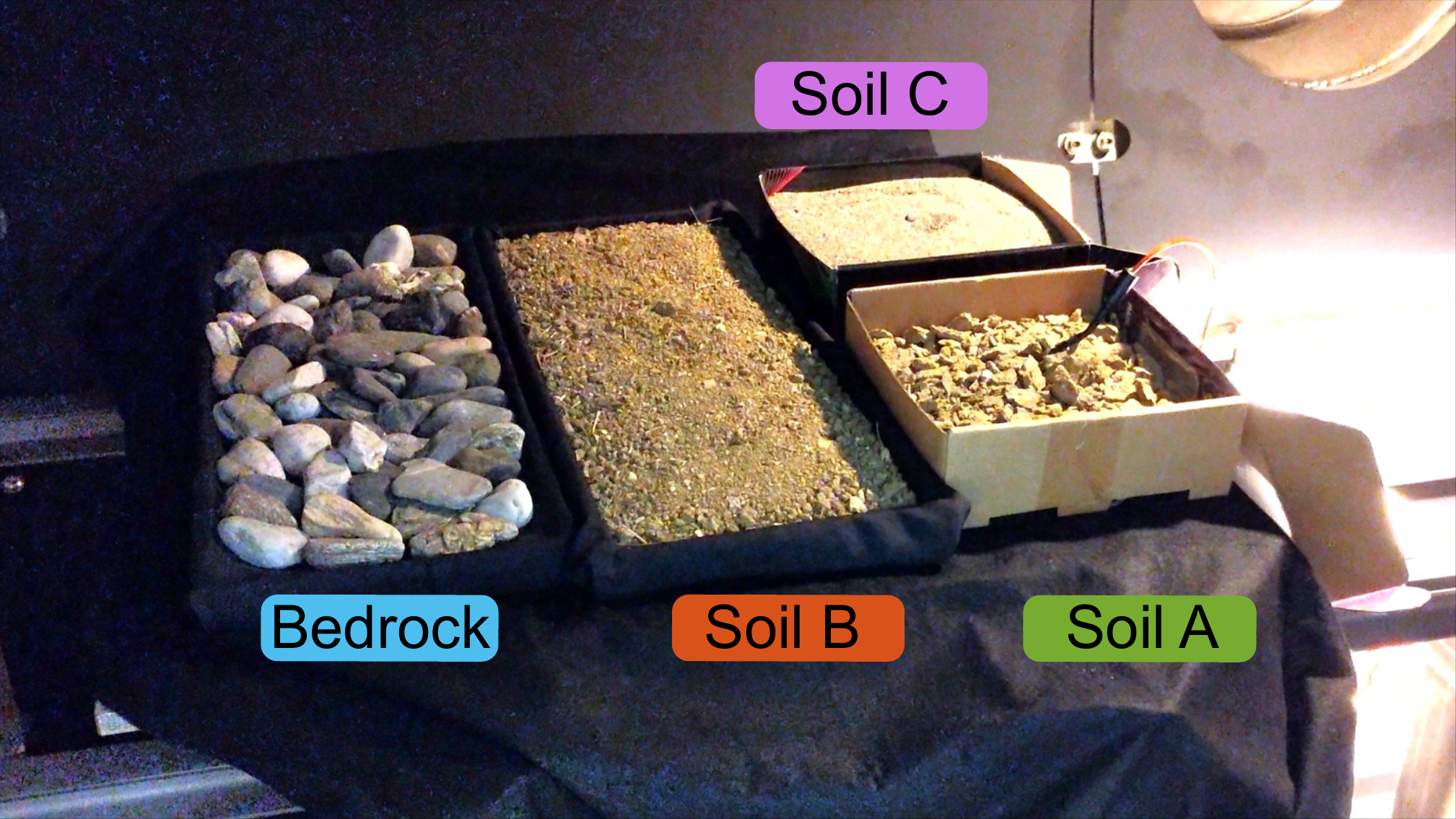}
    \label{fig:chamber-soils}
    \caption{Samples bins of soils of different granularity introduced into the MEC.}
    \label{fig:soils-inside}
\end{figure}

\begin{figure}[t]
    \centering
    \includegraphics[width=0.47\textwidth]{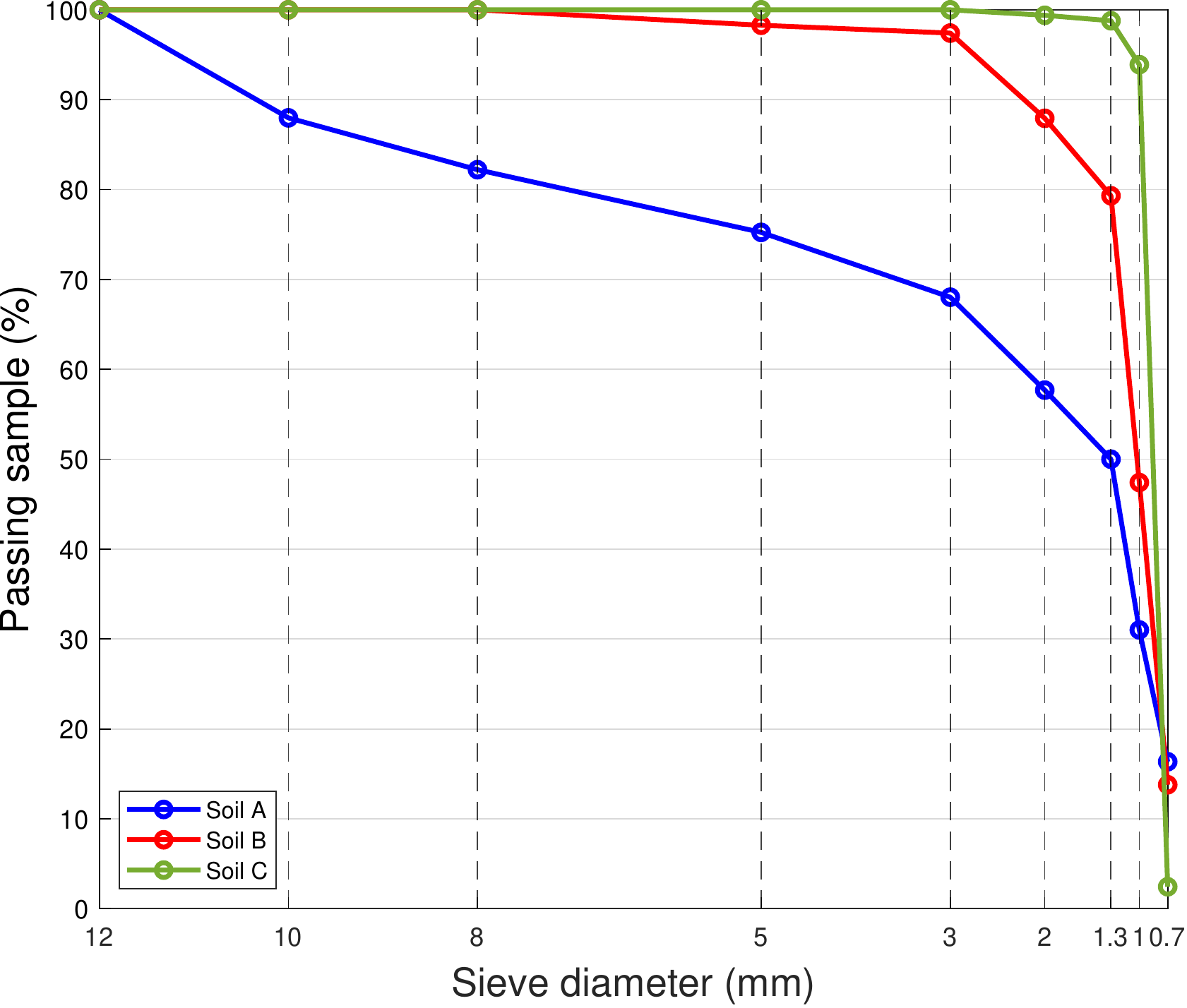}
    \vfill
    \caption{Granularity chart of the soils.}
    \label{fig:granularity-graphs}
\end{figure}

\section{Experimental setup}
\label{sec:experimental-setup}

The measurement system proposed in Section~\ref{sec:framework} has been used to produce representative diurnal cycle results for analyzing the thermal behaviour of four soil samples for corresponding Martian and Earth surface conditions.
This section presents the integration of hardware components to evaluate the proposed system as well as the selection of soil samples.


\subsection{Equipment}

The UMA-Laserlab MEC (see Fig.~\ref{fig:panoramic-martian-chamber}) is a stainless-steel cylinder of $12\:m$ of length and $1.6\:m$ of diameter and viewports on the top and sides~\cite{alvarez2021umachamber}. It is equipped with an inner spot-gridded  thermal jacket or heater which contains a cooling fluid that can reach a temperature in the range of $-72^\circ C$ to $127^\circ C$, at a rate of $1^\circ C / min$. 
The air inside can be pumped out until a pressure of $10^{-4}mbar$ is reached and can be replaced by $CO_{2}$ to simulate the composition of the atmosphere on Mars. It is equipped with vacuum-compliant thermocouple gauges in the center of its core to measure the air temperature. Additionally, the MEC has a stainless steel plate on rails that allow a payload of up to $70\:Kg$. 

The thermal vision camera is a PI-640i by Optris based on uncooled microbolometer technology. 
It is a $320\:g$ Long-Wave Infrared (LWIR) camera that works in the spectral range of $8\mu m$ to $14\mu m$, has a resolution of 640x480 pixels and a germanium optic with a field of view of $60^\circ$x$45^\circ$. It can measure temperatures from  $-20^\circ C$ to $900^\circ C$ with a thermal sensitivity of $0.04^\circ C$. 
We selected this camera due to its high resolution and light weight, making it suitable for mobile and aerial robots. However, this uncooled camera does not provide temperature measurements below $-20^\circ C$, which limited the absolute minimum temperature to which the soil samples could be subjected.  

The thermal camera was connected to a Mini PC Intel NUC with an Intel Core i5 processor of 1.8GHz and 16GB of RAM running the software Optris PIX Connect. We adjusted the focal length of its optic by using a warm body (i.e., a hand) placed on the plate as a reference. The thermal camera geometric and radiometric calibrations were performed and provided by the manufacturer. The sample bins were placed on the plate perpendicular to the thermal camera at a distance of around $1.3\:m$ to have an undistorted view of their surfaces. The plate surface was covered with insulating cardboards and a thick black fabric to avoid IR reflections of the steel.

We designed and developed a viewport adapter (see Fig.~\ref{ffig:martian-viewport}) to remotely make measurements from outside the MEC using the thermal camera. It is composed of a IR window that keeps the inside of the MEC sealed while letting the LWIR range from $8\mu m$ to $14\mu m$ radiation pass through. We chose an anti-reflection coated germanium circular optic model GEW16AR.20 by MKS Instruments due its high mechanical resistance and its ability to withstand abrupt thermal changes. We selected a diameter of $74.9\:mm$ and $5.0\:mm$ of thickness in order to comply with the minimum thickness required to avoid reaching the germanium's fracture strength caused by the pressure differential between the environment and Martian pressure inside the MEC~\cite{yoder2005viewport}. Furthermore, an aluminium toroid frame was crafted to place the unclamped germanium window into the MEC's upper ISO160K compliant viewports.

\subsection{Soil samples}
Four sample bins with soils of different characteristics were selected for the experiments (see Fig.~\ref{fig:soils-inside}). Three of the bins contained granular soils and one contained an example of bedrock. Table~\ref{tab:soil-bins} shows them sorted from highest to lowest mean granularity. We plotted a granularity chart of the granular soils (see Fig.~\ref{fig:granularity-graphs}) by passing them through sieves with grids of different sizes. In terms of homogeneity, Soil C is the most homogeneous, as more than $90\%$ of its grains have a diameter of $0.7\:mm$ to $1\:mm$. It is followed by Soil B, whose grains are mostly concentrated on the size of less than $2.0\:mm$. Finally, Soil A is classified as the most heterogeneous, as it consists of a mixture of several grain sizes. 

\begin{table}[t]
\small
\centering
\caption{\small Sample bins characteristics.}
\begin{tabular}{llll}
Sample & Granularity ($mm$) & Density ($g/ml$) & Bin size ($cm$)  \\
\hline
Bedrock  & 40.0 - 50.0  & 2.94  & 53 x 23 x 3.5  \\
Soil A   & 3.0 - 5.0    & 1.43  & 22 x 22 x 7    \\
Soil B   & 1.3 - 2.0    & 1.40  & 55 x 23 x 3.5  \\
Soil C   & 0.7 - 1.0    & 1.71  & 22 x 22 x 7    \\

\hline
\end{tabular}
\label{tab:soil-bins}
\end{table}

\begin{table}[t]
\small
\centering
\caption{Description of the MEC experiments.}
\begin{tabular}{ccccc}
 & Experiment & Press. ($mbar$) & Subsurf. & $P_{e}$ ($min$)\\
\hline
\multirow{2}{0.8cm}{Pair-1}
& \#1 Earth-like   & 1000  & Soil A  & 296\\
& \#2 Mars-like    & 8     & Soil A  & 297\\
\hline
\multirow{2}{0.8cm}{Pair-2}
& \#3 Earth-like   & 1000  & Soil C  & 320\\
& \#4 Mars-like    & 8     & Soil C  & 360\\
\hline
\end{tabular}
\label{tab:experiment-table}
\end{table}

\begin{table}[t]
\centering
\small
\caption{\small Surface temperatures and heat fluxes of the sampled soils at Earth's and Martian pressures.}
\begin{tabular}[t]{llcccc}
 & & \multicolumn{2}{c}{\footnotesize{Mean Temp.}} &  \multicolumn{1}{c} {\footnotesize{Dev.}} \\ \cmidrule(lr){3-4}\cmidrule(lr){5-5}
& & $T$\textsubscript{init} & $\Delta T$\textsubscript{s}  & $T$\textsubscript{tran}  & $\Delta G$\textsubscript{s}\\
\hline
\multirow{4}{1.2cm}{~\#1 Earth}
& Bedrock   & 24.8  & 53.3  & 1.2 & 280\\
& Soil A    & 24.9  & 51.1  & 1.8 & 280\\
& Soil B    & 25.1  & 51.6  & 0.9 & 274\\
& Soil C    & 24.8  & 51.4  & 1.1 & 275\\
\hline
\multirow{4}{1.2cm}{~\#2 Mars} 
& Bedrock   & 25.0  & 42.0  & 1.7 & 416\\
& Soil A    & 22.7  & 45.7  & 2.5 & 357\\
& Soil B    & 22.4  & 45.5  & 0.4 & 356\\
& Soil C    & 25.0  & 46.3  & 1.0 & 325\\
\hline
\multirow{4}{1.2cm}{~\#3 Earth} 
& Bedrock     & 24.3  & 48.0  & 1.3 & 268\\
& Soil A    & 24.4  & 45.6  & 1.8 & 265\\
& Soil B    & 24.5  & 45.8  & 0.8 & 264\\
& Soil C    & 24.5  & 46.2  & 1.1 & 257\\
\hline
\multirow{4}{1.2cm}{~\#4 Mars} 
& Bedrock     & 25.5  & 38.3  & 1.6 & 399\\
& Soil A    & 24.2  & 41.3  & 2.3 & 337\\
& Soil B    & 24.2  & 41.2  & 1.3 & 337\\
& Soil C    & 24.8  & 43.0  & 0.8 & 310\\
\hline
\end{tabular}
\label{tab:chamber-metrics}
\end{table}

\section{Experiments}
\label{sec:experiments}

This section presents the experiment pairs carried out to evaluate the measurement system proposed in Section \ref{sec:framework}. The recorded thermal images were processed to analyze the soils thermal behaviour and to estimate their thermal inertia.

Two pairs of experiments (Pair-1 and Pair-2) were performed in the MEC on the soils defined in Table~\ref{tab:soil-bins} in order to provide redundant measurements. Table~\ref{tab:experiment-table} summarizes the main characteristics of each experiment. It was only possible to obtain the subsurface temperature of one of the soils per experiment due to MEC connectivity limitations. The subsurface thermocouple gauge was located a depth of $3\:cm$ in soils A and C for Pair-1 and Pair-2, respectively. The experiment pairs consist of Earth representative ($1000\:mbar$) (\#1 and \#3) and Mars representative ($8\:mbar$) (\#2 and \#4) pressures. Besides, or the Mars-like experiments, air with Mars' Carbon Dioxide ($CO_{2}$) atmospheric composition of $95\%$ was introduced into the MEC. On-site near-equatorial environmental measurements performed by Mars Science Laboratory (MSL) showed mean daily air temperatures of around $-50^\circ C$ with approximate amplitudes of $60^\circ C$~\cite{martinez2017modern}. Thus, sinusoidal temperatures of similar amplitude were simulated in diurnal cycles of experimental actuation period $P_e$ by manual input of constant heating and cooling setpoints for MEC actuation.


\begin{figure*}
    \centering
    \subfloat[\label{sfig:exp1-temps-1000mbar-a}Mean temperature]{\includegraphics[height=0.26\textwidth]{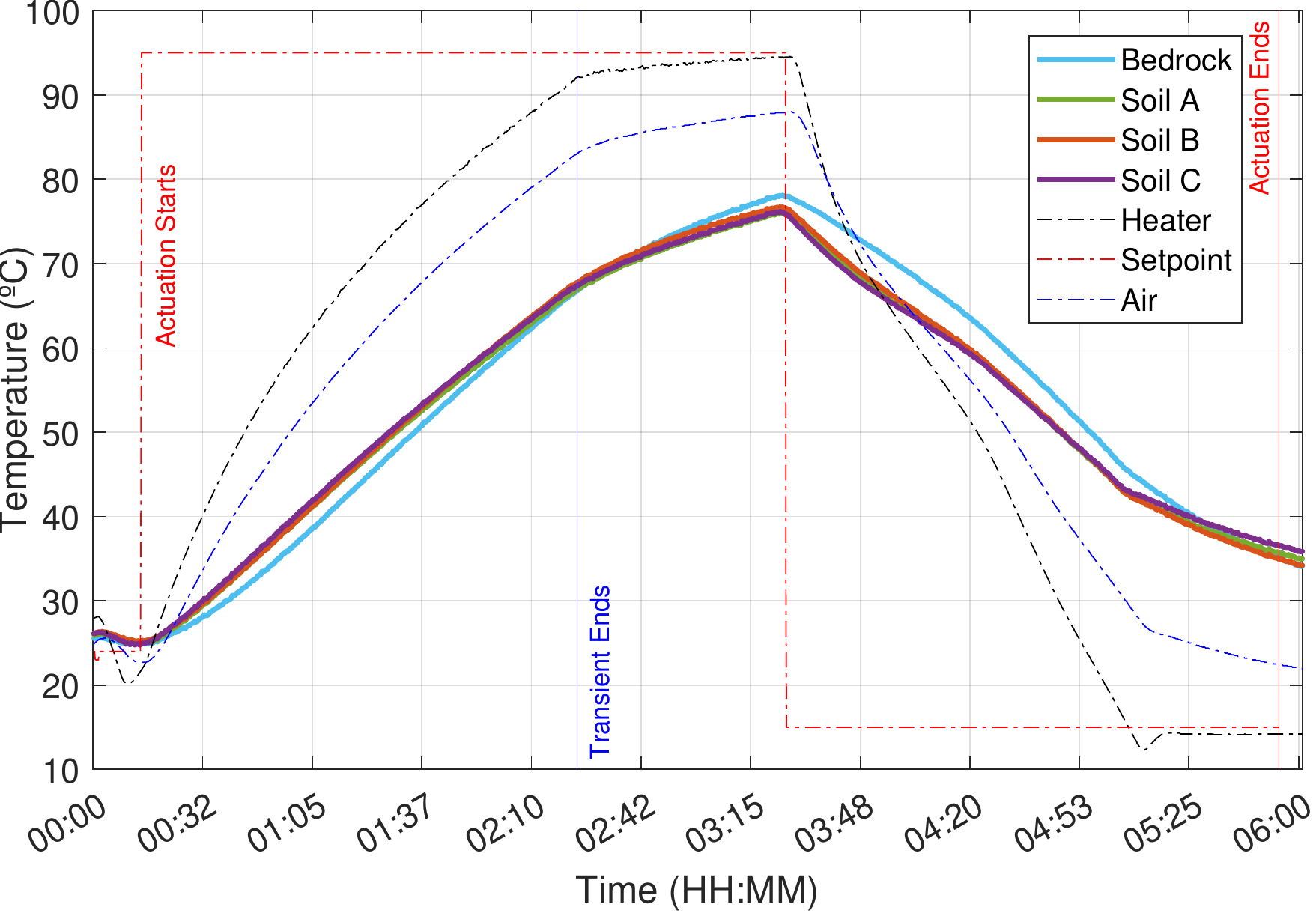}}
    \hspace{1cm}
    \subfloat[\label{sfig:exp1-temps-1000mbar-b}Standard deviation]{\includegraphics[height=0.26\textwidth]{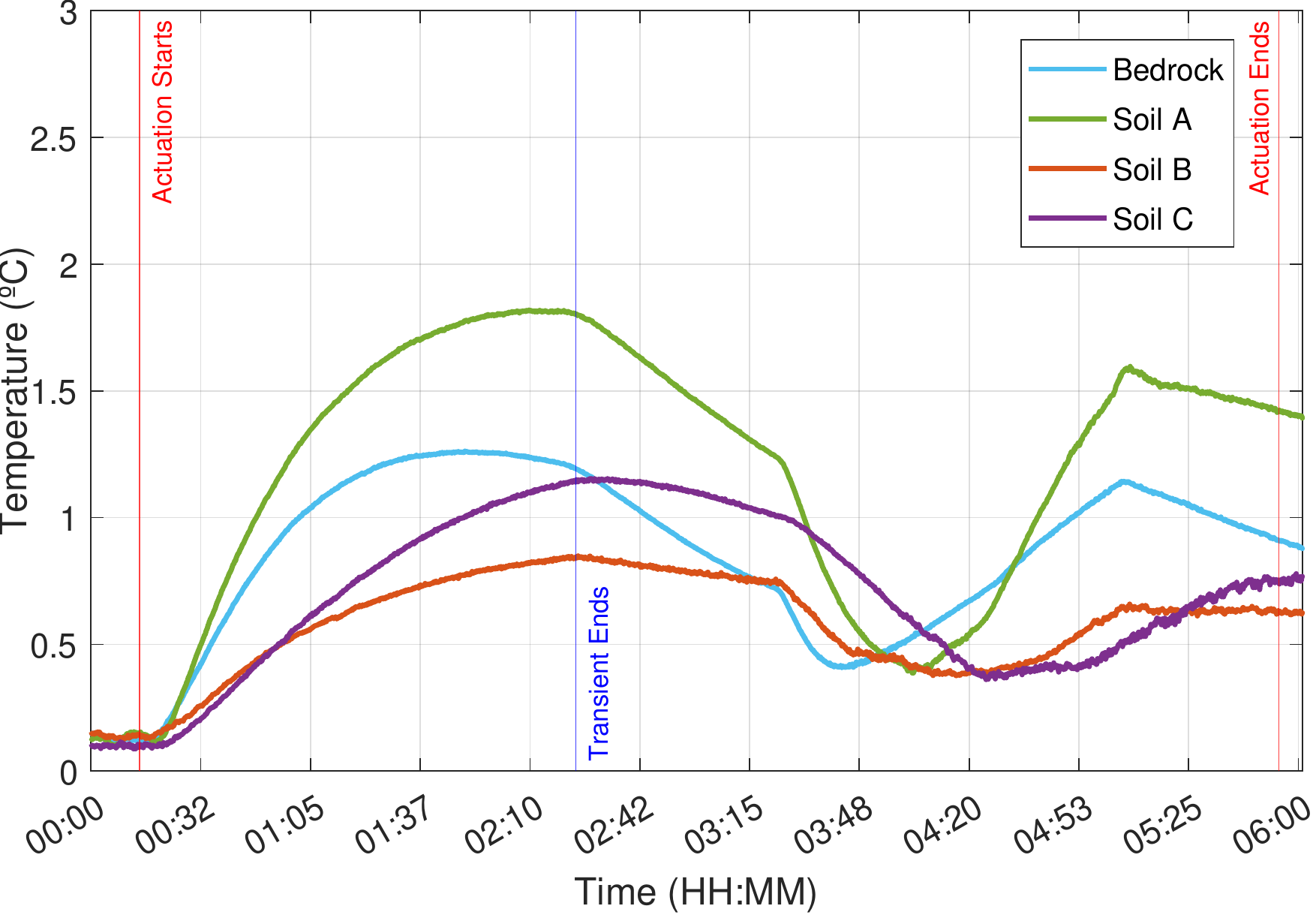}}
    \vfill
    \caption{Diurnal cycle temperatures for the Experiment \#1 at Earth's pressure ($p=1000\:mbar$).}
    \label{fig:exp1-temps-1000mbar}
\vspace{0.3cm}
    \centering
    \subfloat[\label{sfig:exp2-temps-8mbar-a}Mean temperature]{\includegraphics[height=0.26\textwidth]{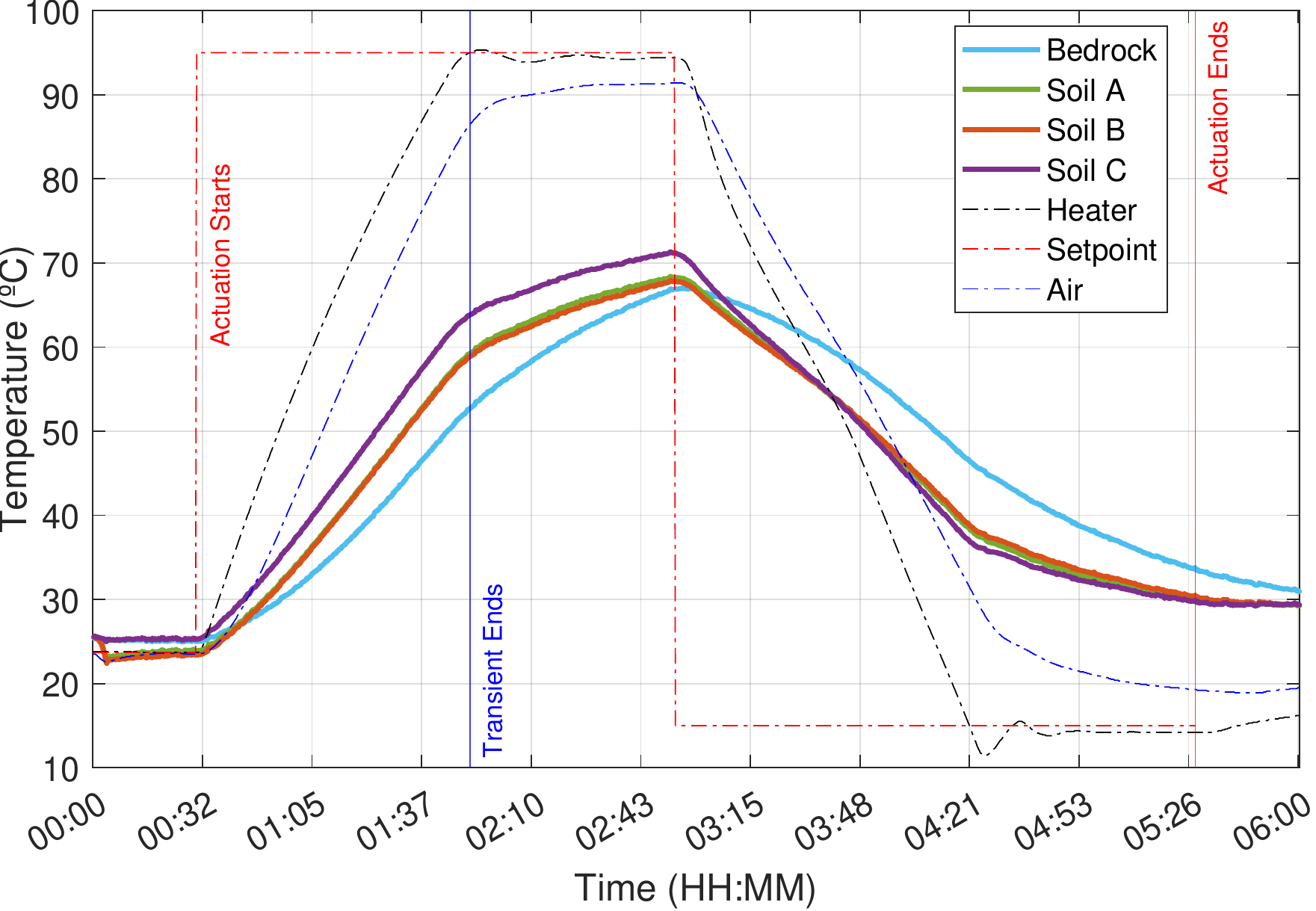}}
    \hspace{1cm}
    \subfloat[\label{sfig:exp2-temps-8mbar-b}Standard deviation]{\includegraphics[height=0.26\textwidth]{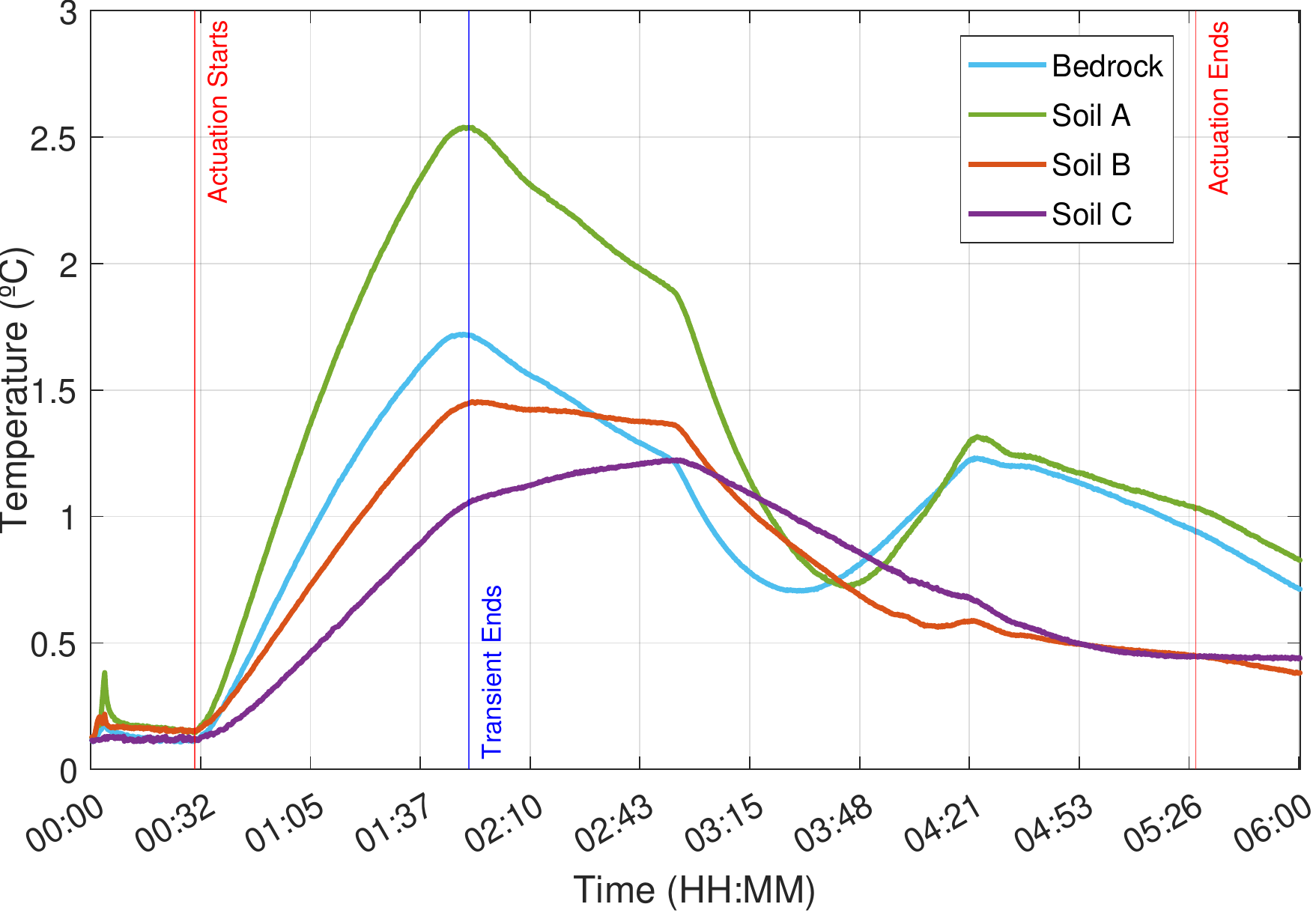}}
    \vfill
    \caption{Diurnal cycle temperatures for the Experiment \#2 at Martian pressure ($p=8\:mbar$).}
    \label{fig:exp2-temps-8mbar}
    \vspace{0.3cm}
    \centering
    \subfloat[\label{sfig:exp3-temps-1000mbar-a}Mean temperature]{\includegraphics[height=0.26\textwidth]{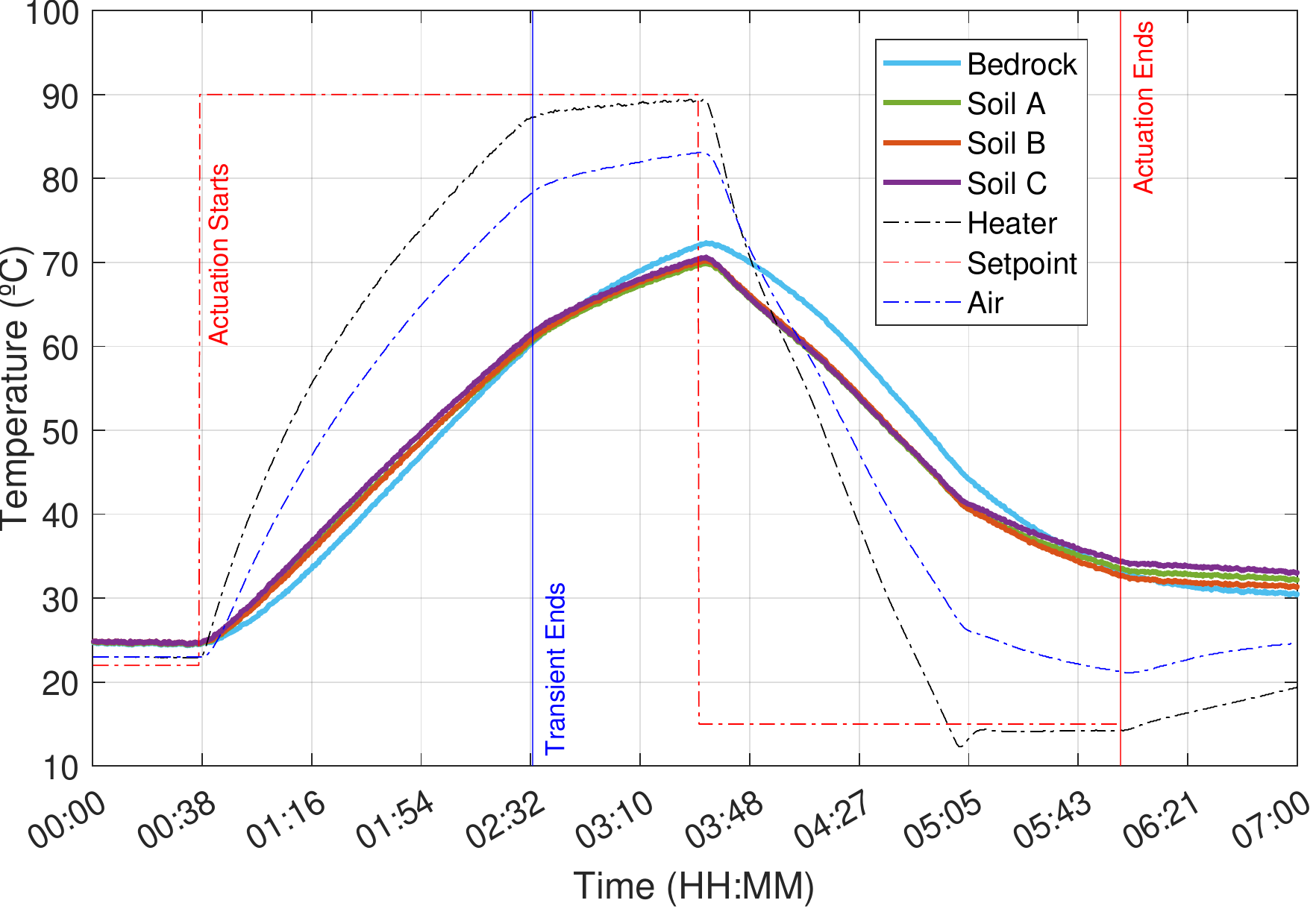}}
    \hspace{1cm}
    \subfloat[\label{sfig:exp3-temps-1000mbar-b}Standard deviation]{\includegraphics[height=0.26\textwidth]{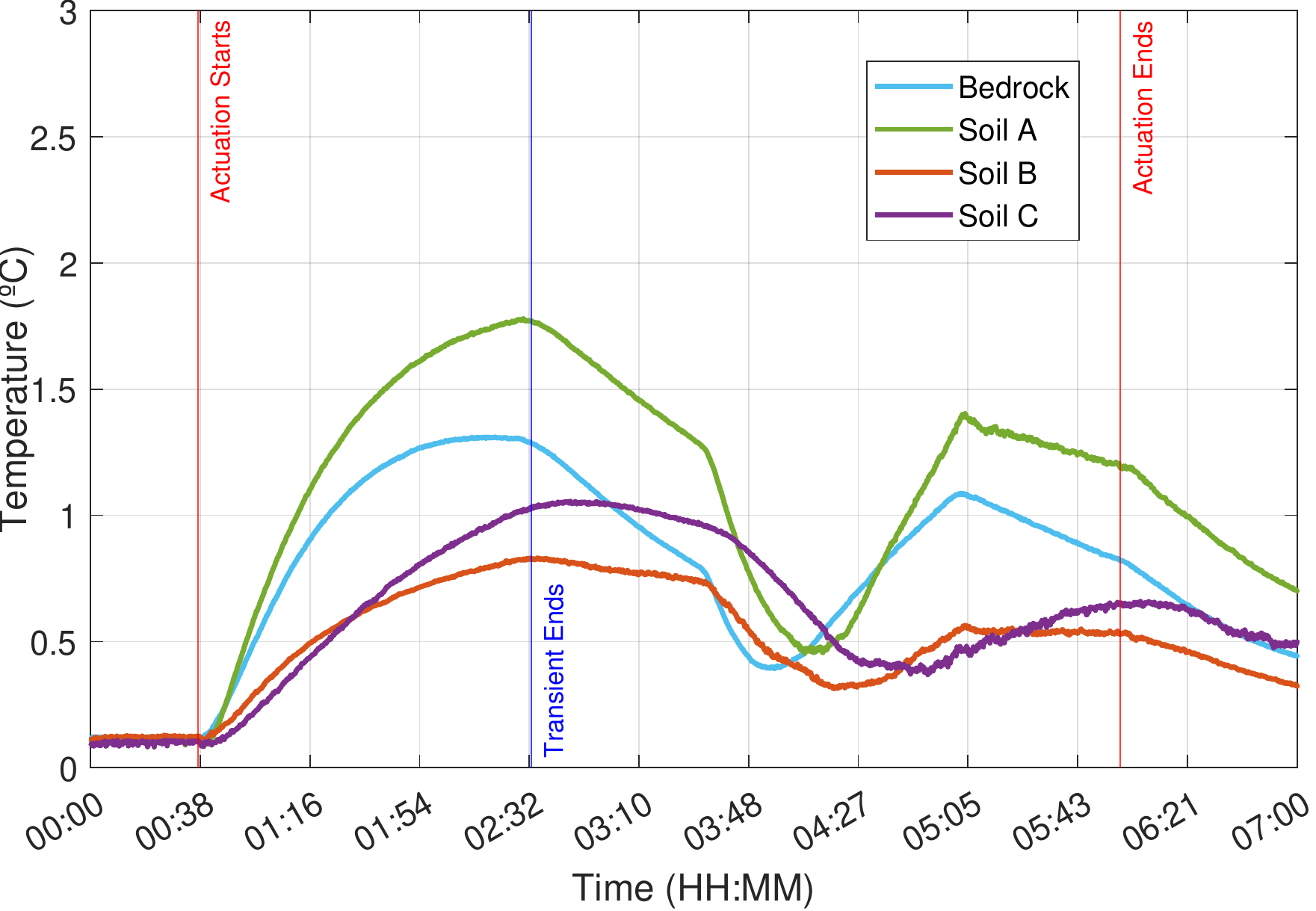}}
    \vfill
    \caption{Diurnal cycle temperatures for the Experiment \#3 at Earth's pressure ($p=1000\:mbar$).}
    \label{fig:exp3-temps-1000mbar}
    \vspace{0.3cm}
    \centering
    \subfloat[\label{sfig:exp4-temps-8mbar-a}Mean temperature]{\includegraphics[height=0.26\textwidth]{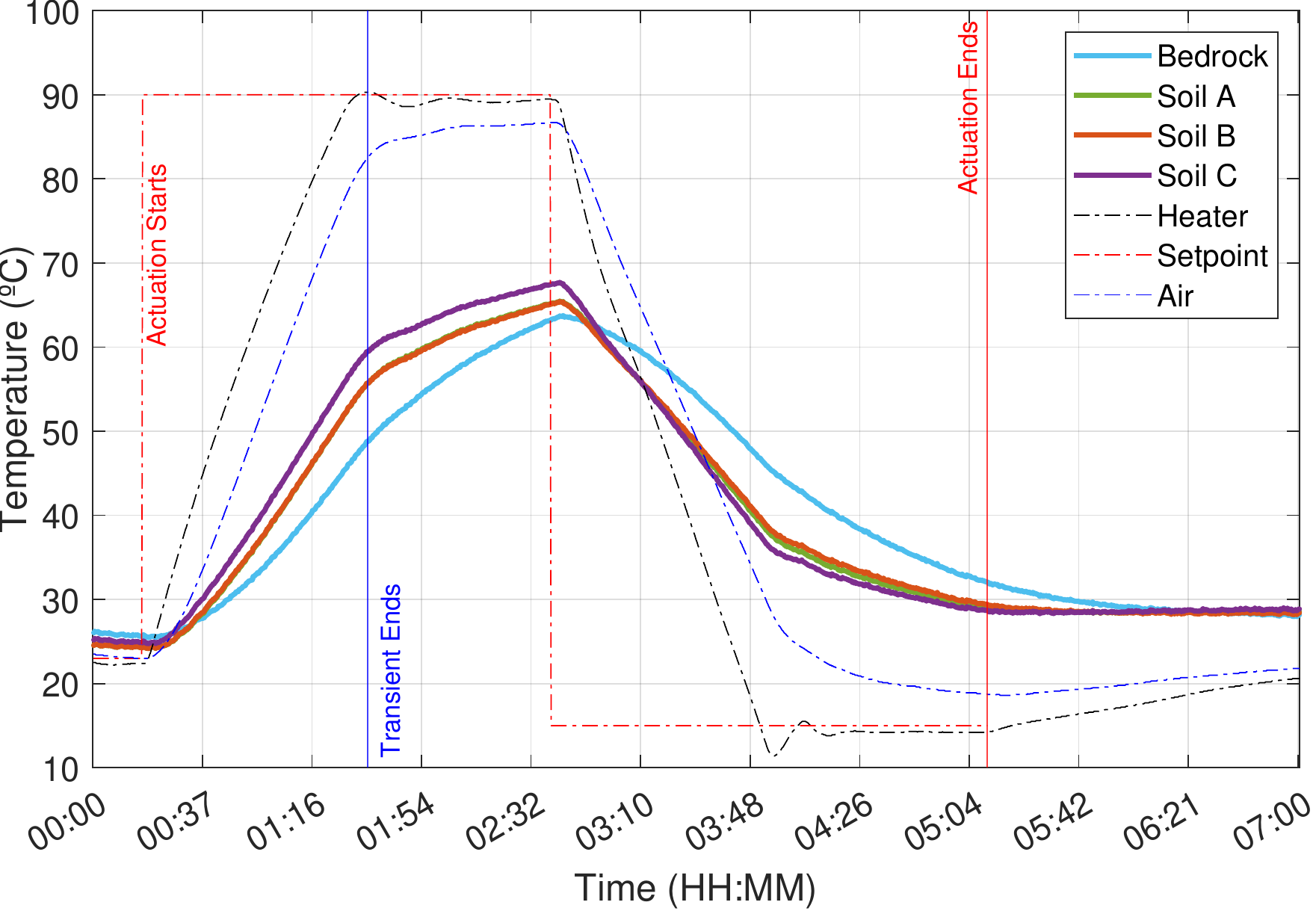}}
    \hspace{1cm}
    \subfloat[\label{sfig:exp4-temps-8mbar-b}Standard deviation]{\includegraphics[height=0.26\textwidth]{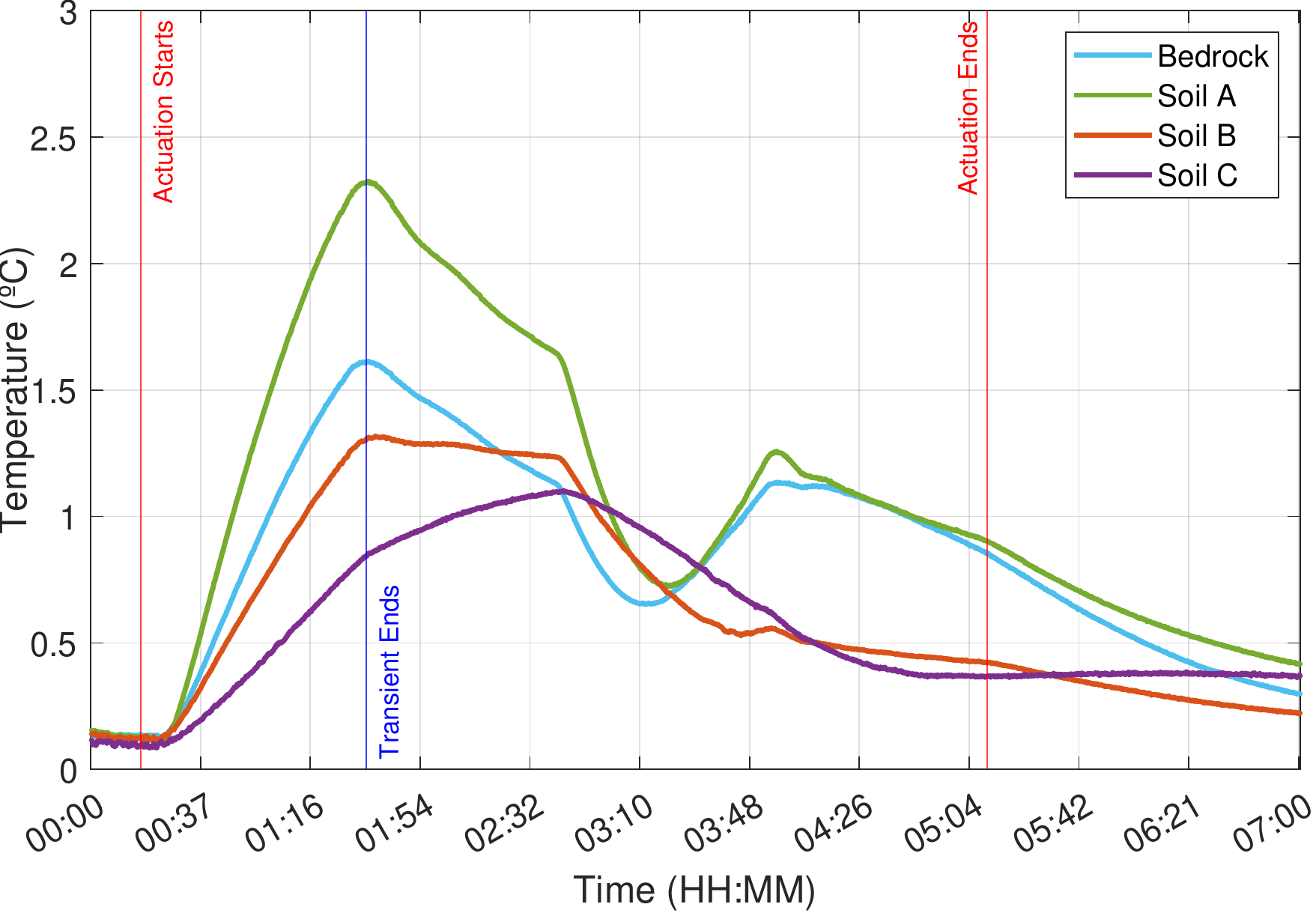}}
    \vfill
    \caption{Diurnal cycle temperatures for the Experiment \#4 at Martian pressure ($p=8\:mbar$).}
    \label{fig:exp4-temps-8mbar}
\end{figure*}

Soil surface temperatures were measured by means of the Optris PI-640i thermal camera. The thermal remote sensing was done as realistically as possible to an actual on-robot implementation, so no prior knowledge of the soils was assumed. Thus, emissivity ($\epsilon$) was considered to be unitary and the albedo ($A$) to be zero for all the soils, according to Kirchhoff's law: $1=A+\epsilon$~\cite{vollmer2021infrared}. Polygonal areas delimiting each soil were defined in the acquired thermal images, where the pixels showing the thermocouple gauge were removed so as not to affect the temperature measurements. 

\begin{figure*}[t]
    \centering
    \subfloat[\label{sfig:soilA-comp-a}Earth's pressure (Experiment~\#1)]{\includegraphics[height=0.26\textwidth]{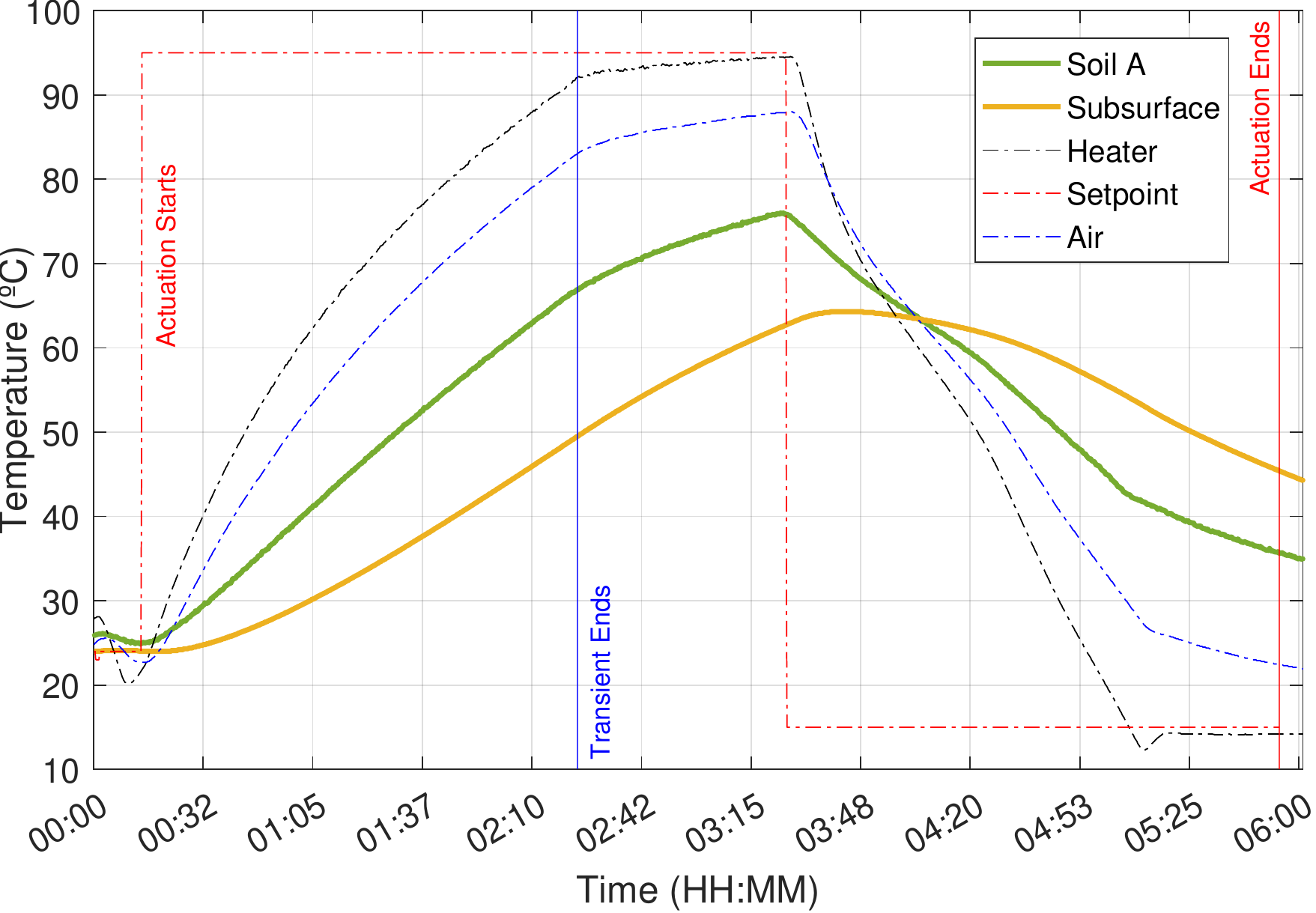}}
    \hspace{1cm}
    \subfloat[\label{sfig:soilA-comp-b}Martian pressure (Experiment~\#2)]{\includegraphics[height=0.26\textwidth]{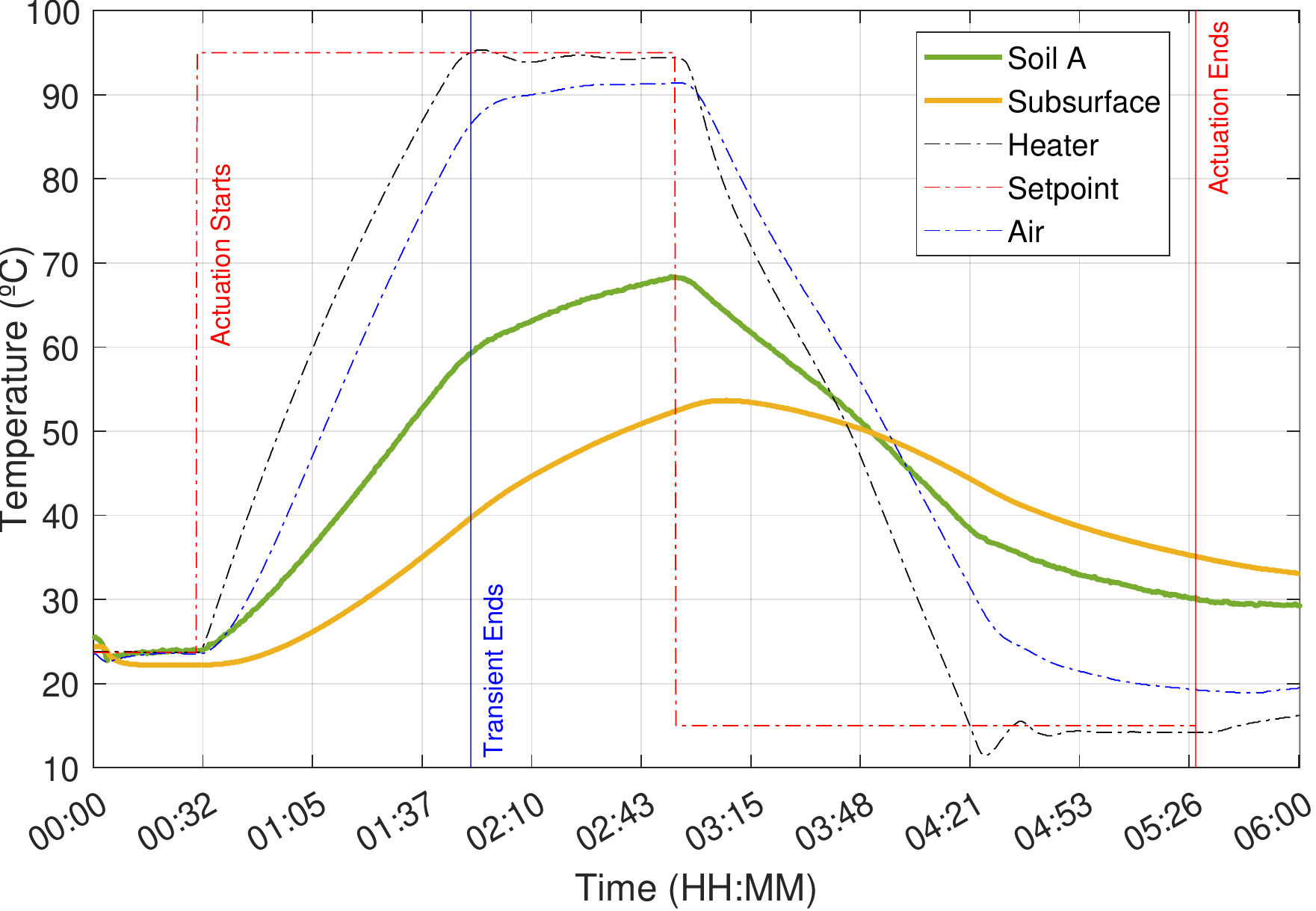}} 
    \vfill   
    \caption{Diurnal cycle surface and subsurface temperatures of Soil A.}
    \label{fig:soilA-comp}
    \vspace{0.3cm}
    \centering
    \subfloat[\label{sfig:soilC-comp-a}Earth's pressure (Experiment~\#3)]{\includegraphics[height=0.26\textwidth]{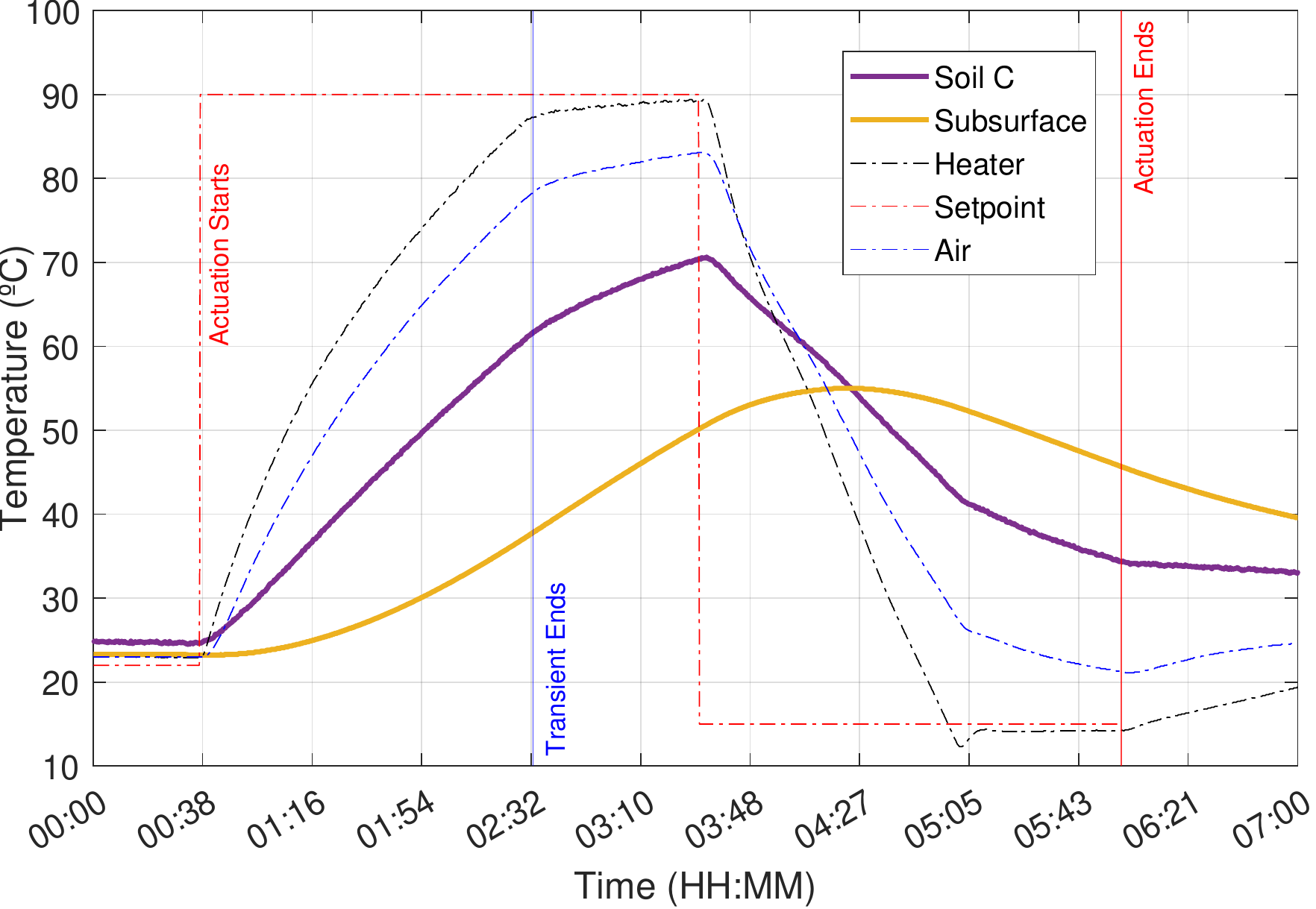}}
    \hspace{1cm}
    \subfloat[\label{sfig:soilC-comp-b}Martian pressure (Experiment~\#4)]{\includegraphics[height=0.26\textwidth]{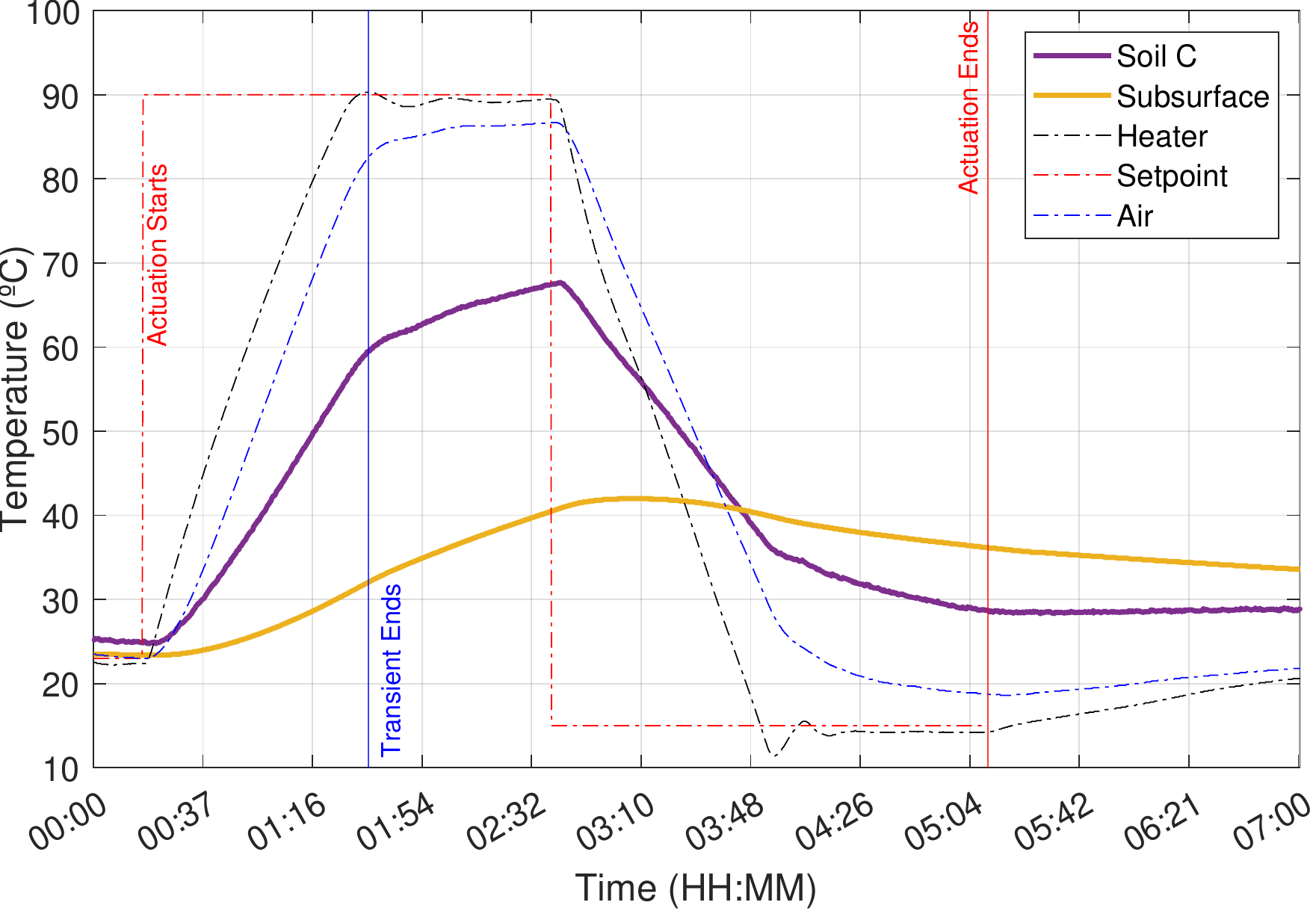}}
    \vfill
    \caption{Diurnal cycle surface and subsurface temperatures of Soil C.}
    \label{fig:soilC-comp}
\end{figure*}

Figures \ref{fig:exp1-temps-1000mbar}-\ref{fig:exp4-temps-8mbar} show the diurnal cycle temperatures for all soil types from experiments \#1-\#4, respectively. Besides, Figs. \ref{fig:soilA-comp} and \ref{fig:soilC-comp} present surface and subsurface temperature readings for soil A (experiments \#1 and \#2) and soil C (experiments \#3 and \#4), respectively. All figures show the heater, setpoint and air temperatures. The transient is assumed to end when an inflection point is reached in the upwards heater temperature response. 
Moreover, Table~\ref{tab:chamber-metrics} presents soil surface mean temperatures for the pixels in the corresponding polygonal area together with standard deviations for the four experiments.  In the table, $\Delta T_{s} = T_{max} - T_{init}$, being $T_{max}$ the maximum mean temperature and $T_{init}$ the mean temperature when actuation starts. $T_{tran}$ specifies the standard deviation temperature of each soil when the actuation transient ends. Net heat fluxes were computed by applying the surface energy budget equation of each sample bin inside the MEC~(\ref{eqn:chamber-flux}) using the soils mean temperatures and the MEC heater temperatures obtained during the experiments. Moreover, the increase of the net heat fluxes during the experiments was calculated: $\Delta G_{s} = G_{max} - G_{init}$, being $G_{max}$ the maximum net heat flux and $G_{init}$ the net heat flux when actuation starts.



\begin{figure}[t]
    \centering
    \includegraphics[width=0.49\textwidth]{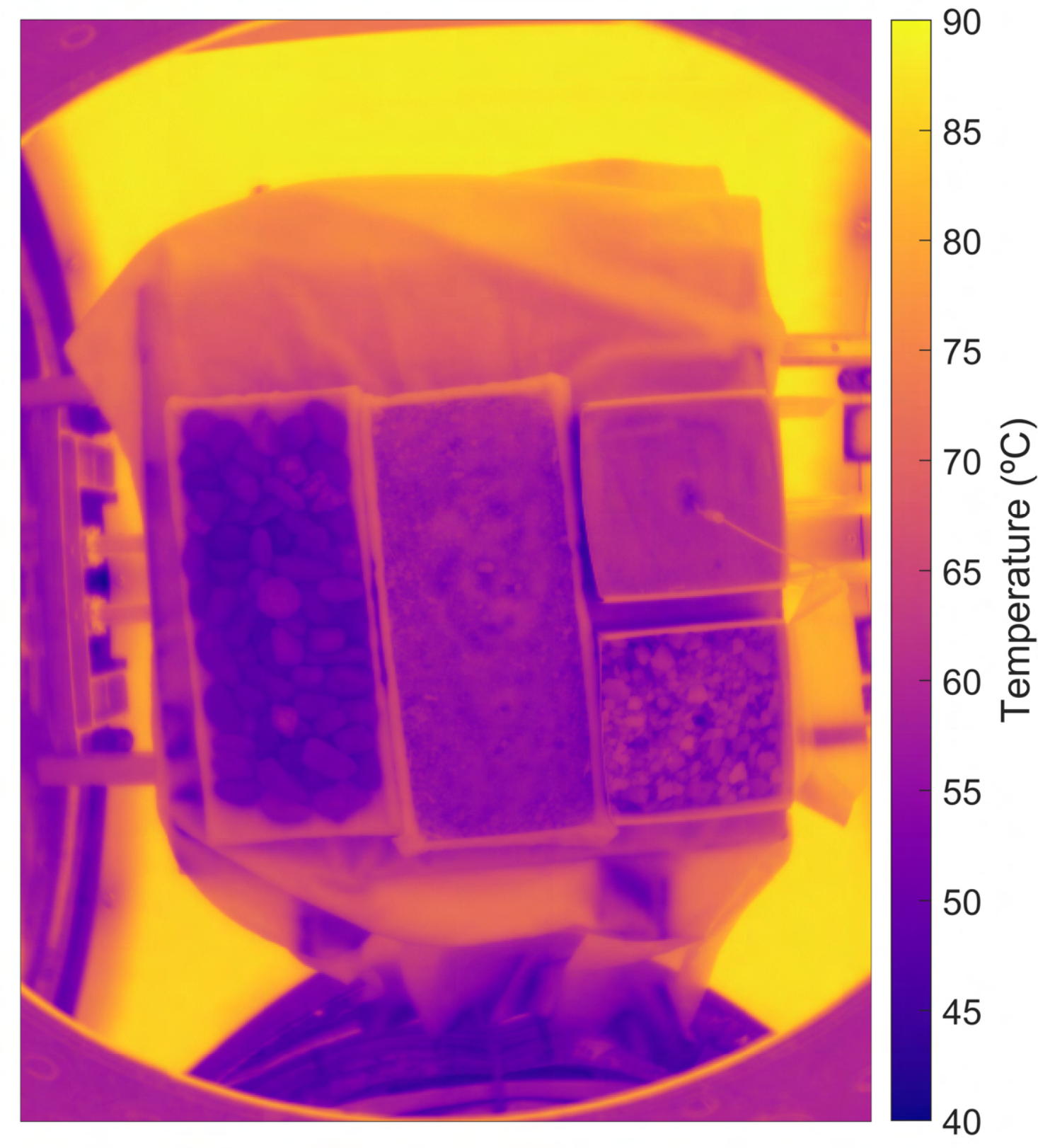}
    \vfill
    \caption{Example of a thermal image of the sample bins during the experiments.}
    \label{fig:thermal-camera-view}
\end{figure}

\subsection{Soils thermal behaviour}

Three features of interest are analyzed from the results obtained in the conducted experiments: the surface mean temperatures, the soils standard deviation curves and the subsurface temperatures.

First, we compare the surface mean temperatures of pair-1 (Figs.~\ref{sfig:exp1-temps-1000mbar-a} and~\ref{sfig:exp2-temps-8mbar-a}). The graphs show that only the bedrock is distinguishable at terrestrial pressure, as all the granular soils present similar temperatures. On the other hand, at Martian pressure, three prominent groups are noticed based on their temperature during the MEC heating; from highest to lowest: (1) Soil C; (2) Soil A and B; and (3) bedrock. Besides, both graphs show a slight temporal delay of the bedrock temperature over the rest of soils. The same analysis can be applied to the graphs of pair-2 (Figs.~\ref{sfig:exp3-temps-1000mbar-a} and~\ref{sfig:exp4-temps-8mbar-a}).

Next, we compare the soils standard deviation curves of pair-1 (Figs.~\ref{sfig:exp1-temps-1000mbar-b} and~\ref{sfig:exp2-temps-8mbar-b}). All the soils can be classified using the standard deviation of the surface temperatures at both Earth's an Mars' pressures from the start of the actuation until the transient ends. This behaviour can be due to their granularity: heterogeneous soils (e.g., Soil A) show higher standard deviation temperatures than more homogeneous soils (e.g., Soil C). This behaviour becomes more evident in the Martian case. A similar analysis can be applied to the graphs of pair-2 (Figs.~\ref{sfig:exp3-temps-1000mbar-b} and~\ref{sfig:exp4-temps-8mbar-b}). 
 
Finally, we compare the mean surface value with the subsurface temperatures of Soil A in pair-1 (Figs.~\ref{sfig:soilA-comp-a} and~\ref{sfig:soilA-comp-b}). In this case, the maximum difference between the surface and subsurface temperature are $11.6^\circ C$ and $14.7^\circ C$  at Earth's and Mar's pressure, respectively; which constitutes an increase of 26.72\%. As for Soil C in pair-2 (Figs.~\ref{sfig:soilC-comp-a} and~\ref{sfig:soilC-comp-b}), for Earth, the maximum difference is $14.4^\circ C$, whereas for Mars-like it is $24.4^\circ C$ ; which is a 69.44\% increase. We can observe that the thermal inertia of both soils increases when the pressure decreases, as it gets more difficult for the heat to be transmitted vertically.

\subsection{Thermal inertia estimation}

\begin{table}[t]
\small
\centering
\caption{ \small Estimated values of thermal inertia for each soil.}
\begin{tabular}[t]{llcccc}
 & & Bedrock & Soil A & Soil B & Soil C \\
\hline
\multirow{4}{0.5cm}{I\textsubscript{sin}}
& ~\#1 Earth  & 309 & 322 & 312 & 314 \\
& ~\#2 Mars & 522 & 411 & 411 & 370 \\
& ~\#3 Earth  & 311 & 323& 320 & 310 \\ 
& ~\#4 Mars & 548 & 431 & 431 & 379 \\ 
\hline
\multirow{4}{0.5cm}{ATI}
& ~\#1 Earth & 79 & 82 & 81 & 81 \\
& ~\#2 Mars  & 100 & 92 & 92 & 90 \\
& ~\#3 Earth & 87 & 92 & 91 & 91 \\
& ~\#4 Mars  & 109 & 101 & 102 & 97  \\
\hline
\end{tabular}
\label{tab:thermal-inertia-est}
\end{table}

We computed estimations of each soil thermal inertia based on the $\Delta T$\textsubscript{s}, $\Delta G$\textsubscript{s} and $P$\textsubscript{e} values obtained during the experiments. For comparison purposes, we used both the ATI~(\ref{eqn:ati}) and the sinusoidal estimation, $I_{sin}$~(\ref{eqn:inertia-estimation}). The estimated values are shown in Table~\ref{tab:thermal-inertia-est}. 


Regarding the sinusoidal estimations, $I_{sin}$, it is observed that thermal inertia increases when pressure decreases. Thus, soils are easier to classify at Martian pressure than at Earth's pressure. Soils with larger particle sizes, e.g., Bedrock, have higher thermal inertia; on the other hand, soils with smaller particles, e.g., Soil C, show lower thermal inertia. The estimated thermal inertia values of the bedrock sample bin are consistent with the on-site thermal inertia obtained by Curiosity for bedrock-dominated surfaces ($\sim$ 350 – 550 $tiu$)~\cite{vasavada2017thermophysical}. As for Soil C, its estimated thermal inertia is similar to surfaces of around $1\:mm$ mean particle size also derived from Curiosity's data ($\sim$ 265 – 375 $tiu$)~\cite{hamilton2014observations}. On the other hand, soils A and B present similar thermal inertia despite having different particle size. In this case,~\cite{jakosky1986thermalmartian} argued that soils with particle sizes from $1\:mm$ to a few centimeters have a constant thermal inertia of about 420 $tiu$. In conclusion, at Earth's pressure, the mean relative difference of the highest inertia soil compared with the lowest inertia soil is 4.20\%; while at Martian pressure the difference is 42.84\%.

As for the ATI estimations, even though the relative differences between soils are consistent, they do not display a significant increase of their absolute values when the pressure decreases. This is mainly due to the fact that ATI does not consider the soils heat fluxes inside the MEC. Thus, thermal inertia estimations using the ATI equation are not adequate enough for this kind of experiments. 

\subsection{Dataset}

During the experiments, we collected a total of 9225 radiometric images. Each image recorded by the thermal camera was saved as a plain text 640x480 matrix with each cell containing the temperature in degrees Celsius. Snapshots of the thermal images were processed to facilitate direct viewing. An example of one of these snapshot is shown in Fig.~\ref{fig:thermal-camera-view}. Finally, spreadsheets were generated with the heaters, air, and subsurface temperatures recorded by the thermocouples. To the authors' knowledge, no similar dataset exists in the literature. A public dataset with the recorded data can be found at Zenodo\footnote{\url{http://doi.org/10.5281/zenodo.7750148}}.

\section{Conclusions and future work}
\label{sec:conclusions}

This article proposes a MEC-based remote thermal measurement system to physically simulate soils thermal behaviour over diurnal cycles under planetary conditions of pressure and atmospheric composition. The obtained results were processed to estimate the thermal inertia values of the soils, which were compared with real on-site estimations performed by rovers of Mars, showing that our measurement system is capable of physically simulating the soil thermal behaviour under Mars' conditions. 

Based on the analysis of the experiments carried out in this paper, we conclude that thermal vision cameras can be useful to remotely assess soils under Martian pressures. This is equally true under Earth's conditions: although relative thermal inertia are less dependant on soils characteristics, measurements of surfaces mean temperatures and standard deviations can potentiality provide information about soils characteristics. Thus, soil classification algorithms based on thermal vision that work on Earth will perform much better on Mars. Our system enables the generation of datasets to train machine learning algorithms focused on on-site terrain segmentation based on soils characteristics using thermal vision. Future space mission could benefit from the use of this kind of algorithms.

Future work may be focused on enhancing the proposed measurement system by decreasing the minimum temperature that can be remotely measured by using cooled thermal cameras. Furthermore, the development of control systems capable of producing adaptable sinusoidal temperature profiles in the MEC could help to increase the realism of the physical simulations.


\section*{Acknowledgements}
This work was supported by the Andalusian Regional Government under the project entitled "Intelligent Multimodal Sensor for Identification of Terramechanic Characteristics in Off-Road Vehicles (IMSITER)" under grant agreement P18-RT-991.


\bibliographystyle{IEEEtran}
\bibliography{IEEEabrv,main}

\end{document}